\documentclass[letter, 12pt, runningheads]{article}

\usepackage{amsmath,amsfonts,bm}

\def\eqref#1{equation~\ref{#1}}

\def\1{\bm{1}}

\def\vmu{{\bm{\mu}}}

\def\vp{{\bm{p}}}

\DeclareMathAlphabet{\mathsfit}{\encodingdefault}{\sfdefault}{m}{sl}
\SetMathAlphabet{\mathsfit}{bold}{\encodingdefault}{\sfdefault}{bx}{n}

\newcommand{\R}{\mathbb{R}}

\DeclareMathOperator*{\argmax}{arg\,max}

\usepackage{amsmath, amsthm}

\newtheorem{theorem}{Theorem}
\newtheorem{proposition}[theorem]{Proposition}
\newtheorem{lemma}[theorem]{Lemma}

\theoremstyle{definition}
\newtheorem{definition}{Definition}

\theoremstyle{remark}

\makeatletter
\newcommand{\inlineanchor}[2]{%
  \ensuremath{#2}%
  \def\@currentlabel{\ensuremath{#2}}%
  \phantomsection
  \ltx@label{#1}%
}
\makeatother

\usepackage[ruled]{algorithm2e}
\usepackage{algpseudocode}
\usepackage{amsmath, amsfonts,amssymb,amsthm}
\usepackage[english]{babel}
\usepackage{booktabs}
\usepackage{bbm}
\usepackage{color}
\usepackage{dsfont}
\usepackage{enumitem}
\usepackage{fancyhdr}
\usepackage{float}
\usepackage[T1]{fontenc}    
\usepackage[colorlinks=true,linkcolor=blue,citecolor=blue,urlcolor=blue]{hyperref}
\usepackage[capitalize,noabbrev]{cleveref}
\usepackage[utf8]{inputenx}
\usepackage[left=1in,top=1in,right=1in,bottom=1in]{geometry}
\usepackage{graphicx}
\usepackage{mathrsfs}
\usepackage{mathtools}
\usepackage{multirow}
\usepackage{nicefrac}       
\usepackage{listings}
\usepackage{parskip}
\usepackage{sidecap}
\usepackage{subcaption}
\usepackage{textcomp}
\usepackage[natbibapa]{apacite}

\usepackage[textwidth=30mm]{todonotes}

\allowdisplaybreaks

\DeclareMathOperator*{\bi}{>}

\newcommand{\debug}[1]{{\color{black}#1}}
\newcommand{\firms}{\debug{\mathrm{F}}}
\newcommand{\firmswom}{\debug{\mathrm{F}\setminus\{m\}}}
\newcommand{\dataset}{\debug{\mathrm{D}}}

\newcommand{\initdata}{\debug{n}}
\newcommand{\nmax}{\debug{n_{\text{max}}}}

\newcommand{\initendow}{\debug{\mathbf{n}}}
\newcommand{\DSdata}{\debug{n}^{\text{DS}}}
\newcommand{\DSendow}{\debug{\mathbf{n}}^{\text{DS}}}
\newcommand{\finaldata}{\debug{N}}
\newcommand{\finalendow}{\debug{\mathbf{N}}}
\newcommand{\decdata}{\debug{\hat{n}}}
\newcommand{\decendow}{\debug{\mathbf{\hat{n}}}}
\newcommand{\pmax}{\debug{p_{\text{max}}}}

\newcommand{\precisionset}{\debug{\mathcal{P}}}
\newcommand{\Poutside}{\debug{P_{\varnothing}}}
\newcommand{\contract}{\debug{\mathbf{Q}}}
\newcommand{\nullcontract}{\debug{\mathbf{I}}}
\newcommand{\fullcontract}{\debug{\mathbf{1}}}
\newcommand{\contractset}{\debug{\mathcal{Q}}}
\newcommand{\socialwelfare}{\debug{\mathrm{SW}}}
\newcommand{\paretoset}{\debug{\mathrm{PO}}}
\newcommand{\gfairule}{\debug{L}}
\newcommand{\onefairule}{\debug{H}}

\allowdisplaybreaks
\begin{document}
\thispagestyle{empty}

\begin{center}
\LARGE{\textbf{Pareto-Optimal and Strategyproof \\ Data-Sharing Between Competitors}}
\end{center}
\medskip

\begin{center}
\textbf{Morgan Colin-Mourre}$^1$\quad \textbf{Felipe Garrido-Lucero}$^2$\quad \textbf{Patrick Loiseau}$^1$
\end{center}

\begin{center}
1. Inria, Fairplay joint team, Palaiseau, France\\
2. IRIT, Université Toulouse Capitole, Toulouse, France
\end{center}
\vspace{1cm}

\begin{abstract}
\noindent We study data-sharing among firms competing in the same market through the quality of their predictive models. We model firms' profit through two main components, precision and market share, where the former depends on the firm's own number of datapoints and the latter depends on the endowments of all firms. We characterize Pareto-optimal contracts as those in which at least one firm receives all of the other firms' datapoints. We then propose a parametric family of data-sharing contracts and show that, under any of them, the relative market shares of all firms remain constant, eliminating the negative effect of sharing data with competitors and equalizing all firms' gains relative to the no-sharing scenario. We further show that one contract within this family satisfies both Pareto-optimality and strategyproofness against firms attempting to under-declare the size of their initial endowment. Finally, using a real-world dataset, we empirically test the gain in profit for all firms under the contract described above, and show that data-sharing is particularly beneficial when the machine learning task is harder or when firms' initial endowments are closer to homogeneous.
\end{abstract}

\section{Introduction}

Modern machine learning and deep learning algorithms increasingly rely on larger and larger amounts of training data, making it harder for firms to possess the required volumes of data on their own. Data-sharing has risen as an established solution in the literature, allowing different data-owners to collaborate in order to train common models \citep{donahue2021model,imakura2021collaborative,hu2019fdml}. Still, a main issue is often ignored in the data-sharing literature, namely that data-owners within the same market are usually competitors selling the same products, and might therefore be reluctant to share. Indeed, successful data-sharing requires data-owners to find each other's data useful, typically requiring them to serve overlapping customers, putting them in direct competition for those same customers. Still, if competitors manage to enter into data-sharing agreements, they could all gain access to more data; in turn, this would result in better models, leading to more clients and higher sales \citep{hagiu2023data}.

In markets such as e-commerce search engines or recommendation systems \citep{ricci2015recommender, zhang2019deep}, firms engage in quality-based competition \citep{taylor2013search}. A firm whose model has a larger predictive power will capture more clients \citep{hagiu2023data}. In such settings, differences in the quality of the models and of the offers produced are the primary factors guiding customer choice, while price-competition is much less prevalent. Search engines usually provide search results to clients at zero cost \citep{taylor2013search}, and in recommendation systems, prices are often a fixed feature of the products being recommended.\footnote{In the real-world data coming from a hotel recommendation system used in Section \ref{sec:experiments}, the price of a hotel is fixed by the hotel itself, and is considered a fixed feature that the recommendation system cannot change.} Data-sharing between competitors might therefore seem counterintuitive, as giving data to a competitor will directly improve their model and help them increase their market share at the firm's expense.

Despite its economic significance, this dilemma has been little studied in the academic literature. \citet{tsoy_data_sharing_btw_competitors} work on a similar question but focus on price-based or quantity-based competition, where additional data reduces production costs or increases products' utilities. On the contrary, we examine a setting where firms are competing on model quality directly and where additional data has an immediate competitive impact, allowing a firm to steal market share from its competitors. In Federated Learning \citep{donahue2021model,kairouz2021advancesFL}, several recent papers have suggested methods for competing firms to collaboratively build machine learning models without exchanging raw data \citep{duopoly_FL}. However, managing competition while respecting privacy-constraints often requires highly sophisticated infrastructures, whether through monetary transfers \citep{CoCoGen, fairness_incentive_scheme_FL} or structural or coalition-based approaches \citep{tan2024fedcompetitors, chen2024free}. In contrast, our work demonstrates that when direct data exchange is possible, a simpler partial data-sharing contract can be sufficient to incentivize cooperation. Finally, our work also connects to the Information Design literature, which has increasingly examined sharing of consumer signals or market states between competitors (\cite{crowdsourcing_market_info,pareto_imp_data_sharing},(\citeyear{coopetition_vs_amazon})); however, in their models, the information shared consists of real-time market signals, while we focus on the sharing of training data prior to market interactions.

\textbf{Contributions.} In this article we consider a set of firms competing in the same market. We formalize the firms' dilemma of sharing data with their competitors by decomposing the profit of a firm into two components, its market share, interpreted as the probability that a new client entering the market chooses to visit the firm, and its precision, related to its ability to close the sale once a client visits the firm. The precision of a firm always increases when engaging in data-sharing, but its market share might decline if competitors gain too much precision themselves. To capture a broad range of data-sharing arrangements, we introduce partial and asymmetric data-sharing, allowing firms to share less than their full dataset and to share different amounts from one another.

The main contributions of the article are the following.

\begin{itemize}[leftmargin = *]
    \item We introduce a novel model of data-sharing under competition where a firm's profit directly depends on the precision of all firms within the market.
    \item We show that in our setting, there always exists a data-sharing contract under which all firms end up better off with respect to the no-sharing setting (\Cref{thm:paretoset}).
    \item We propose a parametric family of data-sharing contracts that ensures all firms obtain an equal gain from data-sharing, eliminating the negative effect of sharing data with competitors, and we prove that a particular contract within this family is both Pareto-optimal (\Cref{thm:properties_fair_rules}) and strategyproof (\Cref{thm:strategyproofness} and Proposition \ref{prop:bound_gain_rounded_alloc}).
    \item We provide an empirical study on real-world data quantifying the gains in profit for firms under the aforementioned contract beyond the theoretical assumptions required to establish our results, and show that data-sharing is particularly beneficial whenever the machine learning task is harder or firms possess similar datasets before sharing.
\end{itemize}

\textbf{Outline}. \Cref{sec:model} introduces our model and main definitions. \Cref{sec:social_welfare_and_pareto_optimality} studies data-sharing contracts maximizing the social welfare of the firms and characterizes the set of Pareto-optimal contracts. \Cref{sec:fairness_and_mechanisms} introduces equal-gain data-sharing contracts, characterizes them through a simple parametrization, and shows that this family contains a Pareto-optimal and strategyproof contract. \Cref{sec:experiments} tests the gains of the aforementioned contract on a real-world dataset, and \Cref{sec:conclusions} concludes the article. Proofs are included in Appendices \ref{app:missing_proofs_A} and \ref{app:missing_proofs_B}, while Appendix \ref{app:discussion_sec_experiments} includes tests on whether the theoretical assumptions required for our results hold in real-life experiments.

\section{Model}\label{sec:model}



We consider a finite set of firms $\firms := \{1,...,m\}$ competing in the same market. Each firm $i \in \firms$ has a dataset $\dataset_i$ representing its past clients, with $\initdata_i := |\dataset_i|$ denoting its size. 
We suppose that all datapoints $x \in \cup_{i\in\firms}\dataset_i$ have been i.i.d sampled from the same distribution, so firms are distinguished only by the size of their dataset. Finally, we denote $\nmax:= \sum_{i \in \firms} \initdata_i$ and $n_{\min}:= \min_{i \in \firms} \initdata_i$.
\medskip

\begin{definition}\label{def:endowments}
Given $\initendow:= (\initdata_1,...,\initdata_m)$ the initial firms endowment and $\DSdata_i$ the number of datapoints received by firm $i \in \firms$ from data-sharing, we denote $\finaldata_i := \initdata_i + \DSdata_i$ its \textbf{final endowment}.
\end{definition}

We model client behavior as a two-stage process. First, upon arrival of a new client to the market, she chooses one firm to visit, with a probability of being chosen that increases with the firm's model precision. Second, the firm observes the client's features and offers a product using a prefixed machine learning model; the client accepts the offer with a probability that increases when the offer is better adapted to her needs. Firms are interested in maximizing their \textbf{profit}, represented by the probability of making a sale. Following the client behavior described above, we decompose a firm's profit into two components: \textbf{market share} and \textbf{precision}. The former refers to the percentage of clients that the firm attracts, while the latter refers to the firm's ability to close a sale upon the arrival of a new client. In other words, market share relates to the probability that a client chooses the firm when entering the market, while precision refers to the probability that, once a client chooses to visit one firm, they will accept an offer from that firm.
We begin by defining firms' precision.
\medskip


\begin{definition}\label{def:precision_functions}
Let $\precisionset:= \{ P : \mathbb{R_+} \to [0,1], \text{ such that } P \text{ is continuous and strictly increasing}\}$ be the set of \textbf{precision functions}. For any fixed $P \in \precisionset$ and firm $i \in \firms$ with final endowment $\finaldata_i$, the \textbf{precision} of firm $i$ is given by $P(\finaldata_i)$.
\end{definition}

Firms' precision is considered to be common across all firms as, upon equal number of datapoints, they are supposed to use equally good ML models achieving the same performance, motivated by the fact that the best performing ML models in the state of the art are typical known by the players within the market. In particular, we assume that different firms will not have significant differences in model precision if they have the same amount of available training data. In economic terms, this corresponds to assuming perfect information on the performance of available ML models and homogeneous production functions.

The monotonicity assumption on the precision functions is connected to assuming that firms datapoints are i.i.d. so, in particular, a larger dataset will represent higher accuracy on prediction. Notice that, since we study firms in competition within the same market, it is natural to assume that all firms have the same kind of clients, and therefore that past observations are indeed independent and identically distributed.

To model market shares, we use a standard model of consumer choice, commonly used in economic and marketing theory \citep{luce2005individual, bell1975market}, which corresponds to the Tullock Contest Function \citep{corchon2010foundations,tullock2001efficient}: given $m$ firms with attractiveness $a_1, \dots, a_m$, the market share of firm $i$ is given by $a_i / \sum_j a_j$. As we study a setting where firms compete on model precision, we consider that clients are more likely to choose a firm with a more precise model, due to, for example, the past experiences of clients and their relatives or common knowledge about the precisions of the different firms. We thus model a firm's attractiveness as a positive function of its precision leading to the following market share definition.
\medskip

\begin{definition}\label{def:market_share}
Given a precision function $P \in \precisionset$ and a firms final endowment $\finalendow = (\finaldata_1,...,$ $\finaldata_m)$, we define the \textbf{market share} of $i \in \firms$ as
\begin{align*}
	m_i(\finalendow) := \frac{P(\finaldata_i)^r}{\sum_{j \in \firms} P(\finaldata_j)^r + \Poutside},
\end{align*}
where $r > 0$ captures how sensitive clients are to differences in the firms' precisions, and $P_\varnothing \geq 0$ represents the global attractiveness of eventual firms outside the data-sharing environment.
\end{definition}


Definition \ref{def:market_share} captures both the positive impact on a firm's market share of receiving additional training data and the corresponding negative impact when its competitors do the same. Indeed, since the precision functions are increasing, notice that $m_i$ is increasing in $\finaldata_i$ and decreasing in $\finaldata_j$, for any $j$ in $\firms\setminus\{i\}$.\footnote{A brief proof of the market share's monotonicity is included in Appendix \ref{app:missing_proofs_A}, Proposition \ref{prop:profit_monotonicity}.} Moreover, Definition \ref{def:market_share} can also be seen from a consumer utility point of view, rooted in discrete choice theory where the standard way of modeling consumers choosing between a discrete set of options in order to maximize their utility is via logit choice probabilities \citep{mcfadden1972conditional, anderson1992discrete}. Indeed, in this setting, if choosing firm $j$ provides a consumer utility $V_j$, firm $i$ is chosen with probability $\exp(V_i) / \sum_j \exp(V_j)$. Therefore, using a standard logarithmic utility $V_i = r\ln(P(\finaldata_i))$, it directly leads to the market share formula given by Definition \ref{def:market_share}.

We introduce next the profit of the firms.
\medskip

\begin{definition}\label{def:profit}
Given a firms final endowment $\finalendow = (\finaldata_1,...,\finaldata_m)$ and a precision function $P\in\precisionset$, we define the \textbf{profit} of firm $i \in \firms$ as $u_i(\finalendow) := m_i(\finalendow)\cdot P(\finaldata_i)$.
\end{definition}

Finally, we define a \textbf{data-sharing instance} as any tuple $\Gamma := (P,\initendow,r,\Poutside),$
with $P \in \precisionset$ a precision function, $\initendow \in \mathbb{N}^m$ a firms initial endowment, $r > 0$ and $\Poutside \geq 0$.

\subsection{Data-sharing contracts}

We model the data sharing among firms through \textit{contracts} which specify the portion of the dataset they agree to share with each of the other agents.
\medskip

\begin{definition}\label{def:data_sharing_contract}
A \textbf{data-sharing contract} consists of any matrix $\contract := (Q_{i,j})_{(i,j)\in m\times m} \subseteq [0, 1]$, such that, for any $i \in \firms$, $Q_{i,i} = 1$. We denote by $\contractset$ the set of all possible data-sharing contracts. Finally, we denote by $\nullcontract$ and $\fullcontract$, respectively, the \textbf{null contract} where no firm shares data, and the \textbf{full contract} where all firms share all their data with all other firms.
\end{definition}

Unlike other models where firms share noisy versions of their datasets \citep{pejo2017together,sun2024profit}, we model each firm's decision as choosing the portion $Q_{i,j}$ of its dataset that firm $i$ is willing to share with firm $j$. Indeed, since all datapoints follow the same prior distribution and firms share several noisy copies at a time, sharing noisy data brings little additional value, as even without knowing the noise distribution, a firm receiving multiple noisy datapoints can estimate it and construct accurate predictors of the underlying non-noisy data \citep{alexander2009deconvolution}. Notice that $Q_{i,j}$ and $Q_{j,i}$ are not required to be equal, allowing for asymmetric sharing among firms.\footnote{Of course, by implementing asymmetric contracts, firms might still share the same number of data points between them, as they might have different initial endowments.} Similarly, a given firm is not forced to share the same amount of data with all the other firms, that is, given $i,j,j' \in \firms$, $Q_{i,j}$ and $Q_{i,j'}$ might take different values.



Given the firms initial endowment $\initendow$ and a contract $\contract\in\contractset$, the final endowment of firm $i \in \firms$ is given by,
\begin{align}\label{eq:final_endowment}
    \finaldata_i(\initendow,\contract) := \initdata_i + \sum\nolimits_{j\in\firms\setminus\{i\}} \lfloor \initdata_j Q_{j,i}\rfloor,
\end{align}
where $\lfloor\cdot\rfloor$ denotes the floor function. When referring to the firms profit, we will abuse of notation and simply denote $u_i(\initendow,\contract)$ when referring to $u_i(\finaldata_1(\initendow,\contract),...,\finaldata_m(\initendow,\contract))$.


Conversely, notice that for any firms' initial and final endowments $\initendow$ and $\finalendow$, such that $\finaldata_i \in [\initdata_i,\nmax]$ for any $i \in \firms$, there always exists at least one contract $\contract\in\contractset$ such that, for any $i \in \firms$, $\finaldata_i = \finaldata_i(\initendow,\contract)$.\footnote{A brief proof of this fact is included in Appendix \ref{app:missing_proofs_A}, Proposition \ref{prop:Q_from_shared_data}.} Since firms' profits are totally determined by their final endowment, throughout the article we will treat both \textbf{contracts and final endowments as equivalent}.

Finally, to keep the analysis simple, we drop the floor function from \Cref{eq:final_endowment} and allow fractional final endowments. This is a mild approximation. Indeed, in large markets like data sharing, final endowments are rarely fractional anyway, and as \Cref{sec:experiments} will show, rounding them to integers barely changes utility.



\section{Social welfare and Pareto-optimality}\label{sec:social_welfare_and_pareto_optimality}

In a fully cooperative setting, a typical technique to determine a desired contract among firms would be to look at the solution that maximizes the social welfare, that is, the sum of the firms profits. 
\medskip

\begin{definition}\label{def:social_welfare}
Let $\initendow$ be a firms initial endowment and $\contract\in\contractset$ a data-sharing contract. We define the \textbf{social welfare} under $(\initendow,\contract)$ by
\begin{align*}
    \socialwelfare(\initendow,\contract) := \sum\nolimits_{i\in\firms} u_i(\initendow,\contract).
\end{align*}
\end{definition}

The monotonicity of the precision functions implies the following result.\footnote{A brief proof of Proposition \ref{thm:social_welfare_maximized_under_full_sharing} is included in Appendix \ref{app:missing_proofs_A}.}
\medskip

\begin{proposition}\label{thm:social_welfare_maximized_under_full_sharing}
For any data-sharing instance $\Gamma$ and $\contract\in\contractset\setminus\{\fullcontract\}$, it holds, $\socialwelfare(\initendow,\fullcontract) > \socialwelfare(\initendow,\contract)$.
\end{proposition}

Unfortunately, since firms see their competitors' market shares increase thanks to their data, full cooperation is highly unlikely to be implementable in practice. Due to this, we aim to characterize the Pareto-frontier of our problem.
\medskip


\begin{definition}
Let $\Gamma$ be a data-sharing instance. A contract $\contract$ is said to be \textbf{Pareto-dominated} by another contract $\contract'$ whenever $u_i(\initendow,\contract') \geq u_i(\initendow,\contract)$ for all $i \in \firms$, with at least one strict inequality. A contract $\contract$ is said to be \textbf{Pareto-optimal} if no other contract Pareto-dominates it. We denote $\paretoset$ the set of all Pareto-optimal contracts.
\end{definition}

The following result gives the full characterization of the set of Pareto-optimal contracts.
\medskip

\begin{theorem}\label{thm:paretoset}
Let $\Gamma$ be a data-sharing instance. It follows,
\begin{align*}
    \paretoset = \{ \contract \in \contractset, \text{ there exists } i \in \firms, Q_{j, i} = 1, \text{ for any } j \in \firms \},
\end{align*}
that is, the set of Pareto-optimal contracts is the set of any contract where at least one firm receives the maximal possible amount of data $\nmax$.
\end{theorem}

\Cref{thm:paretoset} shows that in any situation where no firm is receiving the maximum possible amount of data $\nmax$, it is always possible to find a contract that Pareto-dominates it. More generally, \Cref{thm:paretoset} shows that whenever all firms are able to receive more data, there always exists a way to increase the data received by all the firms such that they all end up better-off (with at least one of them strictly better-off). This result, therefore, strongly encourages data-sharing, even in a competitive setting where all firms only care about maximizing their own profit.

\section{Equal-gain data-sharing} \label{sec:fairness_and_mechanisms}

\Cref{sec:social_welfare_and_pareto_optimality}, and more precisely \Cref{thm:paretoset}, provides a rich family of contracts that can be implemented in a data-sharing setting among competitors. However, real-life firms may have different preferences over which contract in 
$\paretoset$ to choose. Notably, firms do not care only about improving their profits relative to the no-data-sharing setting, but also about the gains being similar across firms. For example, a firm contributing a large amount of data might be reluctant to participate if all its competitors double their sales while its own sales increase by only 10\%. Due to this, this section studies \textit{equal-gain data-sharing contracts}.
\medskip

\begin{definition}\label{def:equal_gain_data_sharing}
Let $\Gamma$ be a data-sharing instance. A contract $\contract \in \contractset$ is said to \textbf{produce equal-gain} if for any two firms $i,j\in\firms$, it holds,
\begin{align*}
\frac{u_i(\initendow,\contract)}{u_i(\initendow,\nullcontract)} = \frac{u_j(\initendow,\contract)}{u_j(\initendow,\nullcontract)},
\end{align*}
where recall that $\nullcontract$ corresponds to the null contract. 
\end{definition}

The family of equal-gain data-sharing contracts is trivially non-empty since the null contract belongs to the family. We characterize next the whole set of equal-gain contracts through a simple parametrization.
\medskip



\begin{proposition}\label{prop:set_fair_rules}
Let $\Gamma$ be a data-sharing instance. The set of equal-gain data-sharing contracts corresponds to the family $(\gfairule^\gamma)_{\gamma\in [0,1]}$, with $\gfairule^\gamma := (\gfairule^\gamma_1, \dots, \gfairule^\gamma_m)  $, for any given $\gamma\in [0,1]$ and $i \in \firms$, verifying,
\begin{align*}
\initdata_i + \gfairule^{\gamma}_i(\initendow) = P^{-1} \biggl( P\Big( \initdata_{j^*} + \gamma \cdot\hspace{-0.3cm}\sum_{k \in\firms\setminus\{j^*\}} \hspace{-0.2cm}\initdata_k \Big)  \cdot\frac{P(\initdata_i)}{P(\initdata_{j^*})} \biggr),
\end{align*}
where $j^*\in\argmax_{j\in\firms} \initdata_j$ can be chosen arbitrarily.
\end{proposition}

Notice that the parametrized family of data-sharing contracts $(\gfairule^\gamma)_{\gamma\in[0,1]}$ does not depend on the two parameters $r$ and $\Poutside$ of the data-sharing instance $\Gamma$. In particular, even if firms disagree on the market's parameters and therefore use different market share functions, the amount of data shared under equal-gain remains unchanged.



We are ready to give one of our main contributions.
\medskip

\begin{theorem} \label{thm:properties_fair_rules}
Let $\Gamma$ be a data-sharing instance and $\gfairule^\gamma$ an equal-gain data-sharing contract, for $\gamma \in [0,1]$. Then, it follows that under $\gfairule^\gamma$, the relative market shares of the firms remain constant, that is, for any $i,j\in\firms$, it follows,
\begin{align*}
    \frac{m_i(\initendow + \gfairule^{\gamma}(\initendow) )}{m_j(\initendow + \gfairule^{\gamma}(\initendow) )} = \frac{m_i(\initendow)}{m_j(\initendow)}.
\end{align*}
In particular, the profit of the firms obtained under the family $(\gfairule^\gamma)_{\gamma\in[0,1]}$ is increasing on $\gamma$. Finally, setting $\onefairule := \gfairule^{\gamma\to 1}$, it follows that $H$ corresponds to a Pareto-optimal contract.
\end{theorem}

Theorem \ref{thm:properties_fair_rules} yields several important remarks. First, notice that equal-gain data-sharing contracts eliminate the negative impact on market shares that arises when competitors receive a larger volume of data. In particular, since a firm's precision is increasing in the size of its final endowment, which increases with $\gamma$, we obtain that taking $\gamma = 1$ maximizes profits, and in particular maximizes social welfare among all equal-gain data-sharing contracts. Intuitively, increasing $\gamma$ up to $1$ increases, for all firms, the amount of received data, at rates that depend on the firms initial endowment $\initendow$. The limit is reached when some firm saturates, that is, when one of the firms attains the maximum final endowment $\nmax$. The firm achieving $\nmax$ is actually the firm(s) with the largest initial endowment, denoted $j^*$, since
\begin{align*}
    \finaldata_{j^*} = \initdata_{j^*} + \onefairule_{j^*}(\initendow) =  P^{-1}\biggl(P(\nmax)\cdot\frac{P(\initdata_{j^*})}{P(\initdata_{j^*})}\biggr) =  \nmax.
\end{align*}
From \Cref{thm:paretoset}, it follows that $\onefairule$ is a Pareto-optimal contract, whereas no other $\gfairule^\gamma$ is.

\subsection{Strategyproofness}

Suppose that firms share their data through a central pool, in which each firm sends a copy of its dataset to a central trusted entity, which then allocates back to each firm the corresponding amount of datapoints prescribed by the data-sharing contract $\contract$.

Under this protocol, firms might be tempted to misreport the amount of data they possess. On the one hand, reporting fewer datapoints decreases the total amount of data received by each firm and, hence, their precision. On the other hand, since a firm only sends a copy of its declared dataset to the central entity, it retains full possession of its true data regardless of what it declares, and can combine it with whatever data it receives back to compute its actual precision. As a result, a firm's own precision is only mildly affected by under-reporting while (potentially) highly deteriorating its competitors' precision. Since a firm's market share depends on its precision relative to that of its competitors, firms might find profitable to under-report their initial endowments.

Since firms are required to send a copy of their declared dataset to the central entity, we suppose they cannot declare more datapoints than they truly possess, as they might fear to be requested to send more data that they actually have. 


For a data-sharing instance $\Gamma$, we denote by $\decdata_i$ the initial endowment that firm $i \in \firms$ declares to possess, and set $\decendow := (\decdata_1,\dots,\decdata_m)$. Given a declared initial endowment $\decendow$, we denote by $\contract(\decendow) \in \mathbb{R}_+^m$ the amounts of additional data received by each firm under a contract $\contract$.
\medskip

\begin{definition}
Let $\Gamma$ be a data-sharing instance. A data-sharing contract $\contract$ is \textbf{strategyproof} if for any $i \in \firms$ and any declared initial endowment $\decendow$, it holds, $$u_i\big( \initendow + \contract( \initdata_i, \decdata_{-i}) \big) \geq u_i\big( \initendow + \contract( \decdata_i, \decdata_{-i}) \big),$$
where $\decendow_{-i} := (\decdata_j)_{j\in\firms\setminus\{i\}}$.
\end{definition}

A contract is strategyproof whenever truthfully declaring one's initial endowment is a best response, regardless of the declared endowments of the other firms. The following result shows that $\onefairule$, in addition to being Pareto-optimal, is also strategyproof.
\medskip

\begin{theorem}\label{thm:strategyproofness}
For any data-sharing instance $\Gamma$ with concave precision function $P$, the data-sharing contract $\onefairule$ is strategyproof.
\end{theorem}

Notice that concavity of the precision function $P$ is a natural assumption as typical machine learning models exhibit diminishing returns to additional data \citep{arnold2018data,chiou2017search}. Additionally, note that Theorem \ref{thm:strategyproofness} holds for continuous final endowments, which, first, is unfeasible in practice and, second, does not ensure that firms might find it profitable to under-report due to the rounding. We address the continuity issue with the following complementary result. We suppose that $n_{\min} \geq 2$.
\medskip

\begin{proposition}\label{prop:bound_gain_rounded_alloc}
Let $\Gamma$ be a data-sharing instance with a concave precision function $P$. Let $\tilde{H}$ denote any integer-rounded version of the continuous equal-gain contract $H$, that is, $\tilde{H}$ is obtained by applying ceiling or floor to each entry of $H$. 
Then, 
for any $i \in \firms$ and any declared initial endowment $\decendow$, the relative utility gain of firm $i$ under rounded allocations is bounded by
\begin{align*}
\frac{u_i\big(\initendow + \tilde{H}(\decdata_i, \decdata_{-i})\big) }{u_i\big(\initendow + \tilde{H}(\initdata_i, \decdata_{-i})\big)}  \le \left( \frac{P(n_{\min})}{P(n_{\min} - 1)} \right)^{4r + 2}\hspace{-0.5cm}.
\end{align*}
\end{proposition}

Proposition \ref{prop:bound_gain_rounded_alloc} upper bounds the utility gain of a firm when under-reporting and an integer-rounded version of the contract $H$ is implemented. Notice that, since $P$ is concave and $n_{\min}\to\infty$, the right-hand side goes to $1$, implying a null profit from under-reporting. Additionally, in non-asymptotic regimes, small values of $n_{\min}$ already guarantee a negligible profit from under-reporting. For example, considering $P(x) = x / (x+1)$ and $n_{\min} = 100$, the maximum gain from under-reporting does not exceed $0.06\%$.

\section{Numerical results}\label{sec:experiments}

This section is dedicated to the implementation of the data-sharing contract $\onefairule$ from \Cref{sec:fairness_and_mechanisms} on a real-dataset. The corresponding code, as well as the values of all hyperparameters computed through cross-validation, are included in the supplementary material of the submission for reproducibility.

\subsection{Expedia dataset}

We use the public dataset from \cite{expedia-hotel-recommendations}, containing more than $38$M random observations from the Expedia website, a search engine to find and book hotels. From the whole dataset, we randomly extracted $500000$ datapoints to use in our experiments. Observations in the dataset contain close to $30$ features each, such as the sub-website used by the customer to search (e.g. Expedia.com, Expedia.co.uk, Expedia.co.jp), the ID of the country the customer is located in, whether the customer was connected from a cellphone, and check-in and check-out dates.

Additionally, based on hotel features, Expedia grouped hotels into $100$ clusters of similar characteristics (price range, star rating, popularity, etc.). As assumed in our model, the Expedia dataset treats price as a feature of the product rather than a parameter set by the firm. Based on these clusters, the dataset records whether a customer booked and, if so, the corresponding cluster.

\subsection{Determining the precision function}

The main technical difficulty in implementing the data-sharing contract $\onefairule$ is computing the precision function, for a fixed machine learning model used by the firms. Indeed, notice that we require knowing the function $P$ at almost all values $n \in \{\min\{\initendow\},\dots,\nmax\}$, which requires training a large number of models.

We consider that firms run a \textit{Random Forest model}, known for its robustness to overfitting on prediction tasks \citep{breiman2001random, hastie2009elements}. After removing all observations where customers did not book any hotel, obtaining a final database with $41000$ samples, we split the remaining dataset into a training dataset $\dataset_{\text{train}}$ and a testing dataset $\dataset_{\text{test}}$, with sizes around $33000$ and $8000$, respectively. Next, to measure the predictive power of the random forest models, we considered different levels of selectivity among customers. A prediction was successful whenever the hotel actually booked by the customer belonged to the top-$k$ predicted clusters output by the model. For example, for $k = 1$, a prediction is successful if and only if the most likely predicted cluster contains the hotel actually booked by the customer, while $k = 3$ requires that the booked hotel belong to one of the three most likely clusters according to the model.


Under this modeling of client behavior, $P(\initdata)$, the probability that a client accepts an offer made by a model trained on $\initdata$ datapoints, can be approximated by the top-$k$ accuracy of such a model on the test dataset. We compute $P_k(\initdata)$ for three different values of $k$ by averaging the top-$k$ accuracy of $20$ models, each trained on $\initdata$ datapoints randomly and independently sampled from the available training data. In practice, we computed $P_k(\initdata)$ at $20$ values of $\initdata$ spread between $100$ and $20000$, and interpolated\footnote{We used a piecewise cubic Hermite spline for interpolation to preserve monotonicity.} between them to obtain the functions $P_k$ for every value of $\initdata$.\footnote{In the experiments, three values of $n$ where required below $100$, corresponding to the regimes $\delta = 0.6$ and $m = 10$, respectively for $k = 1, 3$, and $5$. To calculate those, we simply interpolated the functions $P$ between $0$ and $100$ by fixing $P(0) = 0$, following the idea that a firm with no training data would never have any of its offers accepted.
} Due to the computational burden, $P_k(n)$ for higher values of $n$ were not computed. In particular, as explained in the next section, all created instances considered $\nmax = 20000$.

\begin{figure}[ht] 
    \centering
    \includegraphics[width = 0.9\textwidth]{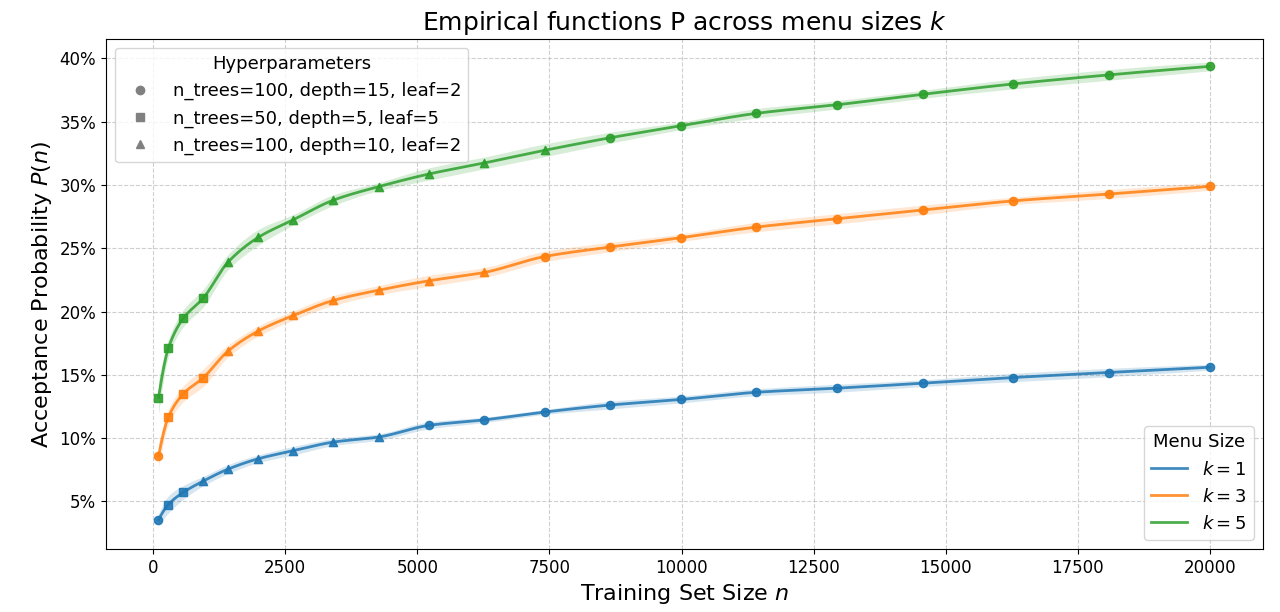}
    \caption{Empirical precision functions $P$ across varying client selectiveness levels $k$. Each curve represents the average top-$k$ test accuracy of a Random Forest model trained on $n$ data points from the Expedia dataset. Accuracies are averaged over 20 random orderings with 3-fold cross-validation for hyperparameters selection. Standard deviations are displayed as intervals around the curve. For each point, the best set of hyperparameters is represented.}
    \label{fig:multiple_P_curves}
\end{figure}

In order to determine the number of trees, the maximum depth of each tree, and the minimum amount of data on each leaf in the random forest models, we considered three different set of parameters (included in Figure \ref{fig:multiple_P_curves}) and performed, for each pair $(k, \initdata)$, a 3-fold cross validation, averaging the scores across $10$ randomly and independently chosen datasets. For any other hyperparameter, their values were set by default.


Figure \ref{fig:multiple_P_curves} includes precision functions for $k \in \{1,3,5\}$. All curves show to be strictly increasing. Regarding concavity, however, the empirical precision functions seem to fail the required assumption for Theorem \ref{thm:strategyproofness} to hold. We further analyze the consequences of the non-concavity of the empirical function $P$ in \Cref{sec:strategyproofness_and_non_concavity}. Finally, we observe a low standard deviation around the $P_k$ curves, guaranteeing the robustness of the contract $\onefairule$ to randomness in the training dataset.

\subsection{Implementing $\onefairule$}

We have looked at the gain in profit for firms across a rich collection of data-sharing instances under the contract $\onefairule$. Since the precision function $P_k$ was computed up to $20000$ datapoints for each $k \in \{1,3,5\}$, we fixed $\nmax = 20000$ as the total number of datapoints available if firms merge their whole datasets. Next, for a given number of firms $m \in \{2,\dots,10\}$, we allocated a value $\initdata_i$ to each $i \in \firms$ given by the following system of equations
\begin{align*}
    &\initdata_i = \delta^{\,i-1}\cdot \initdata_{1}, \text{ for } i \in \{1,\dots,m\}, \text{ and }
    \sum\nolimits_{i\in\firms} \initdata_i = \nmax,    
\end{align*}
where $\delta \in (0,1)$ is a fixed parameter introduced to inject heterogeneity into the firms' initial endowment. In particular, notice that firm $1$ is always allocated the largest number of datapoints. For higher values of $\delta$, the instance is closer to a homogeneous case where all firms have similar initial endowments, while for lower values of $\delta$ the opposite holds.

\begin{figure}[ht] 
    \centering
    \includegraphics[width = 0.9\textwidth]{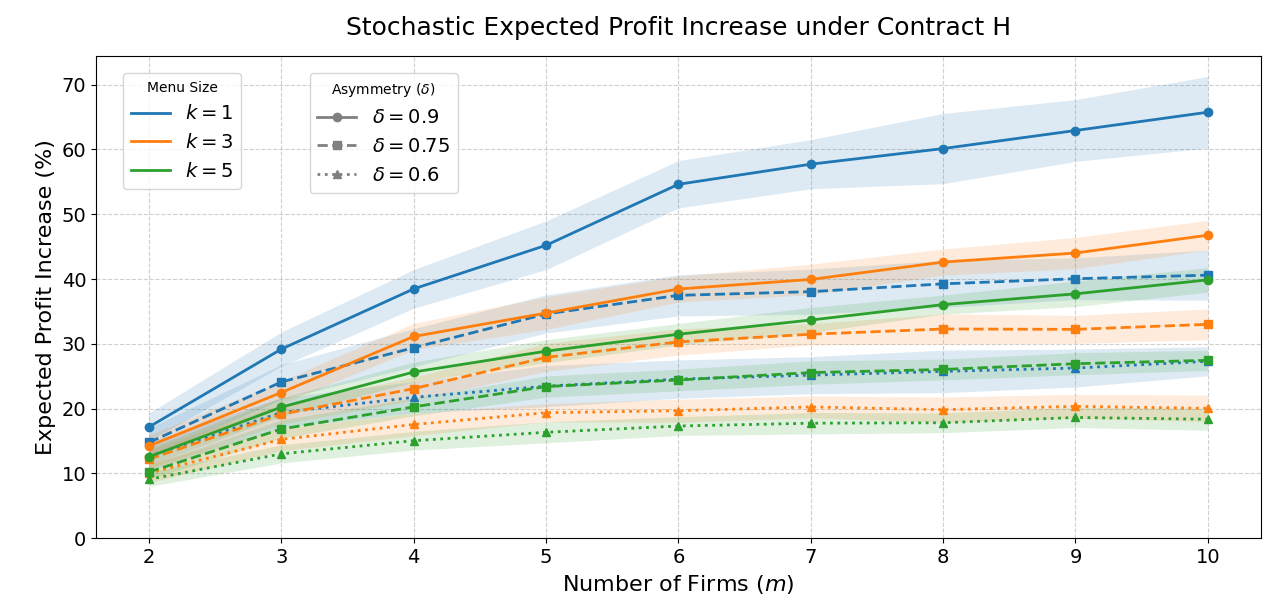}
    \caption{Percentage profits increase of firms under contract $\onefairule$, for $\Poutside=0$ and $r=1$, for different values of client's selectiveness $k$ and market asymmetry $\delta$. The smaller $\delta$, the larger the differences in initial endowments across firms.}
    \label{fig:profit_inc_multinomial}
\end{figure}

Figure \ref{fig:profit_inc_multinomial} shows the gain on profit of firm $1$\footnote{Recall that $\onefairule$ gives the same profit gain with respect to the null contract to all firms.}, in percentage, with respect to the null contract for different values of $k$ and $\delta$, as a function of the number of firms. 
To take into account the uncertainty in the function $P$, all the calculations for the allocations and resulting utilities are made using a stochastic version of $P$, where every time a value $P(n)$ is required for calculations, it is sampled uniformly from the interval $[ \mu(n) - \sigma(n), \mu(n) + \sigma(n)] $, where $\mu(n)$ and $\sigma(n)$ are the empirical mean and standard deviation at point $n$ calculated and displayed in Figure \ref{fig:multiple_P_curves}. Regarding $P^{-1}$, we have considered the average version from Figure \ref{fig:multiple_P_curves} in order to have a monotone function.


Data-sharing shows to be particularly beneficial for smaller values of $k$. Indeed, since a lower value of $k$ represents a harder prediction task, firms obtain a higher benefit from sharing data. Similarly, profits are higher under greater homogeneity of the initial endowments (higher values of $\delta$), as a consequence of the concavity of the precision function. Indeed, since the total amount of data available in the market is fixed at $\nmax$, more homogeneous initial endowments yield, by concavity of $P$, a higher aggregate precision, and consequently a higher aggregate profit. Since the contract $\onefairule$ redistributes firms' data so as to maximize social welfare among equal-gain data-sharing contracts, we observe a higher individual gain.

A third effect can be observed in Figure \ref{fig:profit_inc_multinomial}, namely, the increase in gain with the number of firms. This is an artificial result connected to the previous discussion on homogeneity. As the number of firms increases, since $\nmax$ is fixed, the firms' initial endowment, for a large number of firms, becomes closer and closer to be homogeneous, independently of $k$ and $\delta$.

Overall, the increases are economically significant, ranging from $8\%$ in the worst case, with two firms, a simple prediction task, and high heterogeneity, up to $68\%$ for the most difficult prediction task with several firms and high homogeneity.

Finally, it is important to remark that all results consider fractional final endowments. To complement the theoretical bound in Proposition \ref{prop:bound_gain_rounded_alloc}, we rounded all allocations to compute the gains on integer endowments. The average absolute difference in utility across all instances and firms is roughly $8.4\cdot 10^{-3}$ percentage points, which we consider negligible.

\subsection{Under-reporting incentives on non-concave settings}\label{sec:strategyproofness_and_non_concavity}

The empirical precision functions presented in Figure \ref{fig:multiple_P_curves} are not perfectly concave, as illustrated by our concavity test in Appendix \ref{app:discussion_sec_experiments}. In particular, neither \Cref{thm:strategyproofness} nor Proposition \ref{prop:bound_gain_rounded_alloc} is guaranteed to hold. To complement our results, we have computed the largest possible gains from under-reporting under several different regimes. For the same families of parameters as in Figure \ref{fig:profit_inc_multinomial}, with $m \in \{2,5,10\}$ firms, we considered two strategic scenarios: \textbf{mono-strategic}, where all firms but one report truthfully, and \textbf{bi-strategic}, where all firms but two report truthfully. For each scenario, we considered all possible combinations of firms being strategic or truthful. For mono-strategic cases, we tested all possible deviations $\decdata_i \in [0, \initdata_i]$, while for bi-strategic cases, we considered all possible deviations for both strategic agents in bins of size $10$ (i.e., $\decdata_i \in \{0,10,20,30,...,\initdata_i\}$) for computational reasons. Table \ref{tab:max_gain_no_concavity} includes the maximum percentage utility gain of any strategic agent for each family of parameters under the two considered strategic regimes, compared to declaring truthfully, when implementing the rounded contract $\tilde{H}$ given by the floor version of contract $H$.


The results support the hypothesis that under contract $H$ (or more precisely its rounded version), under-reporting brings poor gains on utility even when the precision function results to be non-concave. Indeed, on mono-strategic settings, the maximum gain achieved across all parameters for any firm is not greater than $0.0380\%$. Moreover, although the gain increases in the bi-strategic setting to $1.4163\%$, the profits remain low.

\begin{table}[h]
\small
\centering
\caption{Maximum gain from under-reporting, under rounded contract $H$, in percentage of the profit when declaring truthfully, given the number of firms $m$, the market asymmetry $\delta$, and the client's selectiveness $k$. In mono-strategic scenarios all firms but one report truthfully, while in bi-strategic scenarios all firms but two report truthfully.}
\label{tab:max_gain_no_concavity}
\begin{tabular}{ccrrrrrr}
\toprule
 \multicolumn{2}{c}{$m$} & \multicolumn{2}{c}{$2$ firms} & \multicolumn{2}{c}{$5$ firms} & \multicolumn{2}{c}{$10$ firms} \\
\midrule
$\delta$ & $k$ & \multicolumn{1}{c}{mono-str} & \multicolumn{1}{c}{bi-str} & \multicolumn{1}{c}{mono-str} & \multicolumn{1}{c}{bi-str} & \multicolumn{1}{c}{mono-str} & \multicolumn{1}{c}{bi-str} \\
\midrule
\multirow{3}{*}{0.9} & 1 & 0.0000\% & 0.1031\% & 0.0000\% & 0.0000\% & 0.0000\% & 0.0000\% \\
& 3 & 0.0000\% & 0.0325\% & 0.0000\% & 0.0005\% & 0.0000\% & 0.0000\% \\
& 5 & 0.0008\% & 0.0376\% & 0.0000\% & 0.0000\% & 0.0000\% & 0.0000\% \\
\midrule
\multirow{3}{*}{0.75} & 1 & 0.0000\% & 0.0497\% & 0.0000\% & 0.0032\% & 0.0000\% & 0.0035\% \\
& 3 & 0.0000\% & 0.0243\% & 0.0000\% & 0.0013\% & 0.0000\% & 0.0032\% \\
& 5 & 0.0012\% & 0.0236\% & 0.0000\% & 0.0006\% & 0.0000\% & 0.0035\% \\
\midrule
\multirow{3}{*}{0.6} & 1 & 0.0037\% & 0.1247\% & 0.0019\% & 0.0048\% & 0.0021\% & 0.0066\% \\
& 3 & 0.0000\% & 0.0401\% & 0.0008\% & 0.0042\% & 0.0034\% & 0.1862\% \\
& 5 & 0.0006\% & 0.0309\% & 0.0000\% & 0.7861\% & 0.0380\% & 1.4163\% \\
\bottomrule
\end{tabular}
\end{table}
\vspace{-0.3cm}

\section{Conclusions}\label{sec:conclusions}

In this paper, we model and analyze data-sharing between firms competing in the same market. Unlike other works in the literature, we study competition on the quality of firms' predictive models. We study the possibilities that arise from partial and asymmetric data-sharing, and prove that solutions always exist under which all firms benefit from cooperating with one another. Moreover, we propose a specific sharing arrangement that eliminates the negative impact on market shares of sharing data with competitors, and show that under this arrangement it is always in firms' best interest to truthfully report their amount of data. Finally, we use real-world data to quantify how much firms could gain from data-sharing, and analyze how market structure influences these gains. Interestingly, similar to \cite{tsoy_data_sharing_btw_competitors}, we find that harder prediction tasks lead to larger gains from data-sharing. Additionally, we find that homogeneous initial endowments increase the gains, possibly explained by the \textit{diminishing returns to data} underlying the concavity of the precision functions, since an additional observations are more valuable to small firms than to larger ones. Homogeneous endowments, in particular, let every firm benefit from this steep, high-return region, whereas a dominant firm's data delivers little extra value to itself while its smaller rivals never receive enough to close the gap.

Despite these contributions, our work has several \textbf{limitations} that suggest directions for future research. First, we model firms as differing only in the number of datapoints they hold, while in practice, firms operating in the same market may hold data that differs in nature or content, not only in size; extending the model to account for heterogeneous data content, rather than heterogeneous data volume alone, is a natural next step. Second, our results rely on a specific market share function, and it would be valuable to test whether they extend to other functional forms. Third, while we show that $\onefairule$ is strategyproof against under-declaring one's initial endowment, we do not rule out over-declaration; although the central-pool implementation described earlier prevents firms from over-declaring, since they are required to physically supply the data they declare, other implementations of the contract might not offer the same protection, and studying strategyproofness against over-declaration remains an open question. Moreover, this guarantee requires for the precision function to be concave, which is not necessarily the case in empirical settings. Fourth, computing the precision function $P$ is computationally costly, since it requires training many machine learning models across a range of dataset sizes, which may limit the practicality of implementing $\onefairule$ at scale. Fifth, our experimental results show substantial gains from data-sharing primarily in markets with some degree of homogeneity among firms' initial endowments, and whether comparable gains persist in more heterogeneous markets remains to be further explored. Finally, although our equal-gain data-sharing contracts balance firms' gains given the current state of the market, we do not study their impact on firms' ability to capture future clients, which could increase relative to the pre-sharing setting.

\textbf{AI use disclosure}. In this work, we used generative AI tools to aid and polish the written text, to retrieve potentially related articles from the literature to our work, and to assist with the Python code developed for the experiments in \Cref{sec:experiments}. All the AI-assisted work was carefully reviewed. No other use of generative AI was done for this work.

\textbf{Acknowledgments}. This work was supported by the ANR LabEx CIMI (grant ANR-11-LABX-0040) within the French State Programme “Investissements d’Avenir.” This work was partially supported by the French National Research Agency (ANR) through the grant ANR-23-CE23-0002 and through the PEPR IA FOUNDRY project (ANR23-PEIA-0003). This work benefited from Hi! PARIS and State funding managed by the French National Research Agency (ANR) under the France 2030 program, reference ANR-23-IACL-0005.

\bibliographystyle{apalike}
\bibliography{literature.bib}

\appendix
\section{Proofs of Section \ref{sec:social_welfare_and_pareto_optimality}}\label{app:missing_proofs_A}

\begin{proposition}\label{prop:profit_monotonicity}
For any data-sharing instance $\Gamma$, it holds that $u_i(\finalendow)$ is strictly increasing on $\finaldata_i$ and strictly decreasing on $\finaldata_j$, for any $j \in\firms\setminus\{i\}$.
\end{proposition}

\begin{proof}
Consider a final endowment $\finalendow$ and two distinct firms $i,j\in\firms$. Consider, for any $x \bi 0$, the functions
\begin{align*}
   f(x) := \frac{x^{r+1}}{x^r + K_1} \text{ and } g_j(x) = \frac{K_2}{x^r + K_3},
\end{align*}
with $K_1, K_2$, and $K_3$ some positive constants. It follows,
\begin{align*}
    f'(x) 
    = \frac{x^{2r} + K_1\cdot (r+1)\cdot x^r } {(x^r+K_1)^2} \text{ and } g_j'(x) = -\frac{K_2\cdot r\cdot x^{r-1}}{(x^r + K_3)^2},
\end{align*}
which are always positive and negative, respectively. Noticing that $u_i(\finalendow) = f(P(\finaldata_i)) = g_j(P(\finaldata_j))$, for well chosen $K_1,K_2$ and $K_3$, and that the precision function $P$ is increasing and non-negative, we conclude the proof.
\end{proof}

\begin{proof}[Proof of Proposition \ref{thm:social_welfare_maximized_under_full_sharing}]
Consider an initial endowment $\initendow$, a contract $\contract\in\contractset\setminus\{\fullcontract\}$ and the corresponding final endowment $\finalendow = \initendow \cdot \contract $. Since $\contract \neq \fullcontract$, there exists a firm $i\in\firms$ such that $Q_{j, i} < 1$ for some $j\in\firms$. In particular, $\finalendow_i < \nmax$. Denote $\pmax := P(\nmax)$. It follows,
\begin{align*}
    \socialwelfare(\initendow, \contract) & = \sum_{i\in\firms} P(\finaldata_i) \cdot m_i(\finalendow) \\
    & < \pmax \sum_{i\in\firms}  m_i(\finalendow)\\
    & = \pmax\cdot \frac{ \sum_{i\in\firms} P(\finaldata_i)^r}{\sum_{j\in\firms}P(\finaldata_j)^r + \Poutside} = \frac{\pmax}{1 + \frac{\Poutside}{ \sum_{j\in\firms}P(\finaldata_j)^r}} .
\end{align*}
On the other hand, 
\begin{align*}
    \socialwelfare(\initendow, \fullcontract) &= \sum_{i\in\firms} P(\finaldata^*_i) \cdot m_i(\finalendow^*) \\
    &= \pmax \sum_{i\in\firms} \frac{p^r_{\text{max}}}{\sum_{j\in\firms}p^r_{\text{max}} + \Poutside} 
    \\
    &= \pmax \cdot \frac{m\cdot p^r_{\text{max}}}{m\cdot p^r_{\text{max}} + \Poutside} 
    = \frac{\pmax}{1 + \frac{\Poutside}{ m\cdot p^r_{\text{max}}}}. 
\end{align*}
We conclude as $\sum_{j\in\firms}P(\finaldata_j)^r < m\cdot p^r_{\text{max}}$.
\end{proof}

\begin{proposition}\label{prop:Q_from_shared_data}
Let $\initendow$ and $\finalendow$ be, respectively, a firms initial and final endowment, where for any $i \in \firms$, $\finaldata_i \in [\initdata_i,\nmax]$. Then, there always exists a contract $\contract \in \contractset$ such that for any firm $i \in \firms$, $\finaldata_i = \initendow\cdot \contract_{\cdot,i}$, where $\contract_{\cdot,i}$ denotes the column $i$ of matrix $\contract$.
\end{proposition}

\begin{proof}
Let $\initendow$ and $\finalendow$ be, respectively, a firms initial and final endowment, such that $\finaldata_i \in [\initdata_i, \nmax] $ for all $i\in\firms$. Let us build a contract $\contract\in\contractset$ such that for any firm $i \in \firms$, $\finaldata_i = \initendow\cdot \contract_{\cdot,i}$. Consider firm $i\in\firms$; if $\finaldata_i = \initdata_i$, simply set $Q_{i, i} = 1$ and $Q_{j, i}=0 $ for all $j\in\firms\setminus\{i\}$. Otherwise, if $\finaldata_i > \initdata_i$, define $S = (S_k)_{k\in\{0, \dots, m\}} $ by:
\begin{align*}
    S_0=0, \quad S_k = \sum_{j=1}^k \initdata_j \mathrm{1}_{j\neq i} \quad \text{for all } k \in \firms.
\end{align*}
As $n_k > 0 $ for all $k\in \firms$, we have $S_k > S_{k-1} $ for all $k\neq i$ (and $S_{i-1} = S_i$). Moreover, $S_m = \sum_{j=1}^m \initdata_j - \initdata_i = \nmax - \initdata_i $. As $\finaldata_i - \initdata_i \in [0, \nmax-\initdata_i] $, there exists a unique $k^* \in\firms \setminus\{i\}$ such that $S_{k^*-1} < \finaldata_i - \initdata_i  \leq S_{k^*}$. As $\contract\in\contractset$, we necessarily have $Q_{i, i} = 1 $. For $j \in\firms \setminus\{i\}$, set:
\begin{align*}
    Q_{j, i} = \begin{cases} 1    &\text{for all } j \leq k^*-1,  \\
    \frac{\finaldata_i - \initdata_i - S_{k^*-1} }{n_{k^*}}   &\text{for } j = k^*, \\ 
    0   &\text{for all } j \leq k^*-1 .
    \end{cases}
\end{align*}
Note that :
\begin{gather*}
    \finaldata_i - \initdata_i > S_{k^*-1} \iff Q_{k^*, i} > 0 \\
    \finaldata_i - \initdata_i \leq S_{k^*} = S_{k^*-1} + \initdata_{k^*} \iff \finaldata_i - \initdata_i - S_{k^*-1} \leq \initdata_{k^*} \iff Q_{k^*, i} \leq 1 
\end{gather*}
which implies that $Q_{j, i} \in [0, 1]$ for all $j\in\firms$. We then have:
\begin{align*}
    \initendow\cdot \contract_{\cdot,i} &= \sum_{j=1}^m \initdata_j Q_{j, i} = \sum_{j=1}^m \initdata_j Q_{j, i} \mathrm{1}_{j\neq i} + \initdata_i Q_{i, i} \\
    &= \sum_{j=1}^{k^*-1} \initdata_j Q_{j, i} \mathrm{1}_{j\neq i} + \initdata_i Q_{i, i} + \initdata_{k^*} Q_{k^*, i} \\
    &= \sum_{j=1}^{k^*-1} \initdata_j \mathrm{1}_{j\neq i} + \initdata_i + \initdata_{k^*} \cdot \frac{\finaldata_i - \initdata_i - S_{k^*-1} }{n_{k^*}} \\
    &= S_{k^*-1} + \initdata_i + \finaldata_i - \initdata_i - S_{k^*-1} = \finaldata_i.
\end{align*}
Proceeding in the same way for all $i\in\firms$ gives a contract $\contract$ such that $\finaldata_i = \initendow\cdot \contract_{\cdot,i}$ for any firm $i \in \firms$.
\end{proof}

\begin{proof}[Proof of Theorem \ref{thm:paretoset}]

\medskip
\noindent{\textbf{Part 1}}: The set of Pareto-optimal contracts is included in $\paretoset$:
\vspace{0.5em}

Let $\initendow \in (0, +\infty)^m $ be an initial data endowment, and consider $\contract \notin \paretoset$ an initial data-sharing contract. We want to construct a new contract $\contract^* $ such that $\contract^*$ is in $\paretoset$ and is a Pareto-improvement over $\contract$. Define $\finalendow = \initendow \cdot \contract $ the final endowment corresponding to contract $\contract$; as $\contract \notin \paretoset$, for all $i \in \firms$, there exists a $j \in \firms$ such that $Q_{i, j} < 1$, which implies that the final endowment of firm $i$ is necessarily smaller than $\nmax$: $\finaldata_i < \nmax $ for all $i \in \firms$. 

On the other hand, any final endowment $\finalendow^*$ such that there exists an $i\in\firms$ with $\finalendow_i^* = \nmax $ necessarily corresponds to a contract where firm $i$ receives all the data from all other firms, that is, $Q_{j, i} = 1$ for all $j\in\firms$. Therefore, any such $\finalendow^*$ would correspond to a contract $\contract^*$ in $\paretoset$. This implies that to find a Pareto-improvement of $\contract$ in $\paretoset$, we simply need to find a new final endowment $\finalendow^*$ with $\finalendow^*_i = \nmax$ for some $i\in\firms$ and that is a Pareto-improvement over $\finalendow$, in the sense that:
\begin{align*}
    u_i(\finalendow^*) \geq u_i(\finalendow) \iff \frac{P(\finaldata^*_i)^{r+1}}{\sum_{j \in \firms} P(\finaldata^*_j)^r + \Poutside}  \geq \frac{P(\finaldata_i)^{r+1}}{\sum_{j \in \firms} P(\finaldata_j)^r + \Poutside} \quad \text{for all } i \in \firms,
\end{align*}
with at least one strict inequality. Once such a $\finalendow^*$ is found, Proposition \ref{prop:Q_from_shared_data} and the above remarks imply that there exists a corresponding $\contract^* \in \paretoset$ such that $\finalendow^* = \initendow \cdot \contract^* $, and this $\contract^*$ would be a Pareto-improvement over $\contract$.

Moreover, we can define a precision vector $\vp = (p_1, \dots, p_m) $ such that $p_i = P(\finaldata_i) $ for all $i \in \firms$. As $P $ is strictly increasing and bijective, finding a Pareto-improvement $\finalendow^*$ of $\finalendow$ is equivalent to finding a precision vector $\vmu$ such that, for all $i \in\firms$,
\begin{align}\label{eq:Ei}
    \frac{\mu_i^{r+1}}{\sum_{j \in \firms} \mu_j^r + \Poutside}  \geq \frac{p_i^{r+1}}{\sum_{j \in \firms} p_j^r + \Poutside} 
\end{align}
with at least one strict inequality, with $\mu_i = P(\nmax)$ for some $i$, and $\mu_j \in [p_j, P(\nmax)] $ for all $j \in\firms$. Indeed, if we can find such a $\vmu$, then setting $ \finaldata_i^* := P^{-1}(\mu_i) $ for all $i\in\firms$ would yield a $\finalendow^*$ satisfying the conditions described above: $\finaldata^*_j \in [\finaldata_j, \nmax] $ for all $j\in\firms$, there exists some $i\in\firms$ such that $\finaldata^*_i = \nmax$ and $\finalendow^*$ is a Pareto-improvement of $\finalendow$.

Without loss of generality, assume that $\finaldata_1 \geq \dots \geq \finaldata_m > 0 \iff p_1 \geq \dots \geq p_m > 0 $. As we assumed that $\contract \notin \paretoset$, we have $\finaldata_i < \nmax$ for all $i\in\firms$ as detailed earlier. Denoting $ \pmax := P(\nmax) $, we therefore have $\pmax > p_i $ for all $i \in \firms$.

We will construct a new precision vector $\mu$ verifying Equations \ref{eq:Ei} and such that $\mu_1 = \pmax$ and $\mu_i \in [p_i, \pmax] $ for all $i\in\firms$ which, as explained above, will determine indirectly a contract $\contract^* \in E$ that will be a Pareto-improvement of $\contract$. Denoting $S = \sum_{j\in\firms} p_j^r + \Poutside  $ and $M = \sum_{j\in \firms \setminus\{m\}} \mu_j^r + \Poutside$, we need, for all $i \in \firms$:
\begin{align*}
    \ref{eq:Ei} \iff \frac{\mu_i^{r+1}}{\sum_{j\in\firms} \mu_j^r + \Poutside} \geq \frac{p_i^{r+1}}{\sum_{j\in\firms} p_j^r + \Poutside} \iff \frac{\mu_i^{r+1}}{M + \mu_m^r} \geq \frac{p_i^{r+1}}{S} 
\end{align*}

When isolating $\mu_m$ in these $m$ inequalities, the conditions for $i\in\firmswom$ lead to upper bounds on $\mu_m$, while the condition for $i=m$ leads to a lower bound:
\begin{align}\label{eq:Em}
    \ref{eq:Ei} &\iff \mu_m \leq \left[ S \left(\frac{\mu_i}{p_i}\right)^{r+1} - M \right]^{1/r} \quad \forall \: i \in \firmswom \\
&\iff \mu_m^{r+1} S - \mu_m^r p_m^{r+1} - Mp_m^{r+1} \geq 0 \notag
\end{align}

For all $ i\in\firmswom $, define $f_i(\mu_1, \dots, \mu_{m-1})$ to be the function determined by \ref{eq:Ei} (recall that $M$ depends on $(\mu_1, \dots, \mu_{m-1})$):
$$  f_i(\mu_1, \dots, \mu_{m-1}) :=  \left[ S \left(\frac{\mu_i}{p_i}\right)^{r+1} - M \right]^{1/r} $$

We look for a vector $\overline{\mu} = (\mu_1, \dots, \mu_{m-1})$ that equalizes all functions $f_i$ for all $i\in \firmswom$, that is:
\begin{align*}
    &f_1(\overline{\mu}) = f_2(\overline{\mu}) = \dots = f_{m-1}(\overline{\mu}) \\
    &\iff \left[ S \left(\frac{\mu_1}{p_1}\right)^{r+1} - M \right]^{1/r} = \left[ S \left(\frac{\mu_2}{p_2}\right)^{r+1} - M \right]^{1/r}  = \dots = \left[ S \left(\frac{\mu_{m-1}}{p_{m-1}}\right)^{r+1} - M \right]^{1/r}   \\
    &\iff \frac{\mu_1}{p_1} = \frac{\mu_2}{p_2} = \dots = \frac{\mu_{m-1}}{p_{m-1}}  \\
    &\iff \mu_i = \mu_1 \frac{p_i}{p_1} \quad \forall \: i\in\firmswom
\end{align*}

We can see that such a $\overline{\mu}$ is entirely determined by the value taken by $\mu_1$. If fixing $\mu_1 = \pmax$, we therefore obtain $\overline{\mu} = \left( \pmax, \pmax\frac{p_2}{p_1}, \dots, \pmax \frac{p_{m-1}}{p_1} \right)$. We look for a $\mu_m$ such that $\mu = (\overline{\mu}, \mu_m )$ satisfies conditions \ref{eq:Ei} for all $i \in \firms$. By definition of $\overline{\mu}$, the first coordinate $\mu_1$ of $\mu$ will be equal to $\pmax$, so the corresponding $\finaldata^*_1 := P(\mu_1)$ will be equal to $\nmax$, as required to obtain a $\finalendow^*$ corresponding to a contract $\contract^* \in E$. Moreover, all other coordinates of $\overline{\mu}$ will be lower than $\pmax$, as $\overline{\mu}_i = \pmax \frac{p_i}{p_1}$ and $p_i \leq p_1$ for all $i\in\firmswom$.

Defining $A_m := f_1(\overline{\mu}) = \dots = f_{m-1}(\overline{\mu})$, the point $A := (\overline{\mu}, A_m)$ trivially satisfies \ref{eq:Ei} for $i \in \firmswom$, but not necessarily \ref{eq:Em}. We therefore define another point $B := (\overline{\mu}, B_m)$, with $B_m$ such that the function defined by \ref{eq:Em} is equal to $0$ at point $B$. Rigorously, with $\mu_i = \pmax \frac{p_i}{p_1}$ for all $i \in \firmswom$, we have:
\begin{align*}
    M &= \Poutside + \sum_{j\in\firmswom} \mu_j^r  = \Poutside + \left(\frac{\pmax}{p_1}\right)^r \sum_{j\in\firmswom} p_j^r 
    = \Poutside + \left(\frac{\pmax}{p_1}\right)^r (S - p_m^r - \Poutside) \\
    &= \left(\frac{\pmax}{p_1}\right)^r (S - p_m^r) - \left( \left(\frac{\pmax}{p_1}\right)^r - 1 \right) \Poutside
\end{align*}

which implies that $A_m := f_1(\overline{\mu})$ is given by:
\begin{align*}
    A_m &= \left[ S \left( \frac{\pmax}{p_1}\right)^{r+1} - M \right]^{1/r} \\
    &= \left[ S \left( \frac{\pmax}{p_1}\right)^{r+1} - \left(\frac{\pmax}{p_1}\right)^r (S - p_m^r) + \left( \left(\frac{\pmax}{p_1}\right)^r - 1 \right) \Poutside \right]^{1/r} \\[5pt]
    &= \left[ \left(\frac{\pmax}{p_1}\right)^r \left( S \left( \frac{\pmax}{p_1} - 1 \right) + p_m^r \right) + \left( \left(\frac{\pmax}{p_1}\right)^r - 1 \right) \Poutside \right]^{1/r}
\end{align*}

Since $\pmax \ge p_1$, we have $\left(\frac{\pmax}{p_1}\right)^r \ge 1$, which ensures that all terms inside the bracket are strictly positive. 

Finally, define the function $f_m : (P(0), \pmax]^m \longrightarrow \R$ determined by Equation \ref{eq:Em}:
\begin{align*}
    f_m(\mu_1, \dots, \mu_{m-1}, \mu_m) = \mu_m^{r+1} S - \mu_m^r p_m^{r+1} - Mp_m^{r+1}.
\end{align*}

We have $ \ref{eq:Em} \iff f_m(\mu_1, \dots, \mu_{m-1}, \mu_m) \geq 0$. We can now define a new function $F \colon [p_m, \pmax] \to \R$, corresponding to $f_m$ when $(\mu_1, \dots, \mu_{m-1})$ is fixed to $\overline{\mu}$:  
\begin{align*}
    F(u) &:= f_m(\overline{\mu}, u) = u^{r+1} S - u^r p_m^{r+1} - M p_m^{r+1} \\
    &= u^{r+1} S - u^r p_m^{r+1} - \left(\frac{\pmax}{p_1}\right)^r (S - p_m^r) p_m^{r+1} + \left( \left(\frac{\pmax}{p_1}\right)^r - 1 \right) \Poutside p_m^{r+1}
\end{align*}
We would like to define $B_m$ as the point where $F(B_m) = 0$; to do so, we first need to prove that it exists and is unique. Taking the derivative of $F$ gives:
\begin{align*}
    F'(u) &= (r+1) u^r S - ru^{r-1} p_m^{r+1} = u^{r-1} \left[ r\left( uS - p_m^{r+1} \right) +uS \right] \\
    &\geq u^{r-1} \left[ r\left( p_m S - p_m^{r+1} \right) +p_mS \right] = u^{r-1} p_m \left[ r\left( S - p_m^r \right) + S \right] > 0
\end{align*}
as $u \geq p_m > 0$ and $S > p_m^r > 0$. $F$ is therefore strictly increasing on $[p_m, \pmax]$. We now need to show that $F(p_m) < 0$ and $F(\pmax) > 0$. Firstly,
\begin{align*}
    F(p_m) &= p_m^{r+1} S - p_m^{r+1}p_m^r - \left(\frac{\pmax}{p_1}\right)^r (S - p_m^r) p_m^{r+1} + \left( \left(\frac{\pmax}{p_1}\right)^r - 1 \right) \Poutside p_m^{r+1} \\
    &= p_m^{r+1} (S - p_m^r) \left[ 1 - \left(\frac{\pmax}{p_1}\right)^r \right] - p_m^{r+1} \Poutside \left[ 1 - \left(\frac{\pmax}{p_1}\right)^r \right] \\[5pt]
    &= p_m^{r+1} \left[ 1 - \left(\frac{\pmax}{p_1}\right)^r \right] (S - p_m^r - \Poutside) < 0
\end{align*}
as $S > p_m^r + \Poutside $ and $\pmax > p_1 \implies \left(\frac{\pmax}{p_1}\right)^r > 1 $. Secondly,
\begin{align*}
    F(\pmax) &= \pmax^{r+1} S - \pmax^r p_m^{r+1} - \left(\frac{\pmax}{p_1}\right)^r (S - p_m^r) p_m^{r+1} + \left( \left(\frac{\pmax}{p_1}\right)^r - 1 \right) \Poutside p_m^{r+1} \\
    &\geq \pmax^{r+1} S - \pmax^r p_m^{r+1} - \left(\frac{\pmax}{p_1}\right)^r (S - p_m^r) p_m^{r+1} \\
    &= \pmax^r \left[ S \left(\pmax - \frac{p_m^{r+1}}{p_1^r} \right) - p_m^{r+1} \left(1 - \frac{p_m^{r}}{p_1^r} \right) \right]
\end{align*}
and since $\pmax > p_1 \geq p_m > 0$, $\pmax - \frac{p_m^{r+1}}{p_1^r} > p_1 - \frac{p_m^{r+1}}{p_1^r} = p_1 \left( 1 - \frac{p_m^r}{p_1^r} \cdot \frac{p_m}{p_1} \right) \ge p_1 \left( 1 - \frac{p_m^r}{p_1^r} \right) \geq 0$. Moreover, using $S \geq p_m^r \implies Sp_1 \geq p_m^{r+1}$ we obtain:
\begin{align*}
    F(\pmax) &> \pmax^r \left[ S p_1 \left( 1 - \frac{p_m^r}{p_1^r} \right) - p_m^{r+1} \left(1 - \frac{p_m^{r}}{p_1^r} \right) \right] 
    > \pmax^r \left( 1 - \frac{p_m^r}{p_1^r} \right) \left[Sp_1 - p_m^{r+1} \right] \geq 0
\end{align*}
To recapitulate, $F$ is continuous and strictly increasing on $[p_m, \pmax]$, $F(p_m) < 0$ and $F(\pmax) > 0$. Therefore, by the Intermediate Value Theorem, there exists a solution $u^* \in (p_m, \pmax)$ such that $F(u^*)=0$, and by strict monotonicity, this solution is unique. We can therefore define $B_m \in [p_m, \pmax]$ to be the unique solution such that $F(B_m) = 0$. Note that $B_m < \pmax$ as $F(\pmax) > 0 = F(B_m) $, and $B := (\overline{\mu}, B_m)$ satisfies \ref{eq:Em} by construction.

We have left to show that $B := (\overline{\mu}, B_m)$ satisfies Equations \ref{eq:Ei} for $i \in \firmswom$, which requires showing that $B_m \leq f_i(\overline{\mu}) = A_m$ for all $i \in \firmswom$; we will additionally show that these inequalities are strict, (as we need at least one strict inequality to obtain a Pareto-improvement). To show that $B_m < A_m$, we simply have to show that $F(A_m) > 0 = F(B_m)$, which is sufficient as $F$ is strictly increasing. We have:
\begin{align*}
    F(A_m) &= A_m^{r+1} S - A_m^r p_m^{r+1} - M p_m^{r+1}.
\end{align*}
Moreover, by definition we have:
\begin{align*}
    A_m = \left[ S \left( \frac{\pmax}{p_1}\right)^{r+1} - M \right]^{1/r} \iff M = S \left( \frac{\pmax}{p_1}\right)^{r+1} - A_m^r
\end{align*}
Substituting this into $F(A_m)$ yields:
\begin{align*}
    F(A_m) &= A_m^{r+1} S - A_m^r p_m^{r+1} - \left( S \left( \frac{\pmax}{p_1}\right)^{r+1} - A_m^r \right) p_m^{r+1}
    = S \left[ A_m^{r+1} - \left( \frac{\pmax}{p_1} p_m \right)^{r+1} \right].
\end{align*}
Therefore, we have:
\begin{align*}
    F&(A_m) > 0 \iff A_m^r > \left( \frac{\pmax}{p_1} p_m \right)^{r} \\
    &\iff\left(\frac{\pmax}{p_1}\right)^r \left( S \left( \frac{\pmax}{p_1} - 1 \right) + p_m^r \right) + \left( \left(\frac{\pmax}{p_1}\right)^r - 1 \right) \Poutside  > \left( \frac{\pmax}{p_1} p_m \right)^{r} \\
    &\iff \left( \frac{\pmax}{p_1} p_m \right)^r + \left(\frac{\pmax}{p_1}\right)^r S \left( \frac{\pmax}{p_1} - 1 \right) + \left( \left(\frac{\pmax}{p_1}\right)^r - 1 \right) \Poutside > \left( \frac{\pmax}{p_1} p_m \right)^r \\
    &\iff \left(\frac{\pmax}{p_1}\right)^r S \left( \frac{\pmax}{p_1} - 1 \right) + \left( \left(\frac{\pmax}{p_1}\right)^r - 1 \right) \Poutside > 0
\end{align*}
which always holds, as $\pmax > p_1$. In conclusion, $F(A_m) > 0 = F(B_m)$ always holds.

This implies that $B_m < A_m = f_i(\overline{\mu})$ holds for all $i\in\firmswom$, and therefore that $B = (\overline{\mu}, B_m)$ satisfies Inequalities \ref{eq:Ei} strictly, and $B_i \in [p_i, \pmax]$, for all $i \in \firms$. Moreover, note that we defined $B$ in such a way that $B_1 = \pmax$. Defining $\finalendow^*_i := P^{-1}(B_i) $ for all $i\in\firms$ therefore gives a new final endowment $\finalendow^*$ that is a Pareto-improvement over $\finalendow$ and such that $\finaldata_i \in [n_i, \nmax]  $ for all $i \in\firms$, and with $\finaldata^*_1 = \nmax$, which concludes the proof that the set of Pareto-optimal contracts is in $\paretoset$.

\medskip
\noindent\textbf{Part 2: Every contract in $\paretoset$ is Pareto-optimal:}
\vspace{0.5em}

Let $\initendow$ be an initial endowment; we want to prove that any contract $\contract\in\paretoset$ is Pareto-optimal, i.e. that it is impossible to build a Pareto-improvement of $\contract$ belonging to $\contractset$. As discussed at the beginning of this proof, any contract $\contract\in\paretoset$ corresponds to a final endowment $\finalendow = \initendow \cdot \contract $ with at least one $i\in\firms$ such that $\finaldata_i = \nmax$. Therefore, we will prove that for any such $\finalendow$, it is impossible to build an $\finalendow^*$ that is a Pareto-improvement over $\finalendow$.

To do so, we prove a slightly more general result; that for any endowment $\finalendow$, a Pareto-improvement $\finalendow^*$ of $\finalendow$ necessarily verifies $\finaldata_i^* > \finaldata_i$ for all $i \in\firms$. Therefore, if there exists an $i \in \firms$ such that firm $i$ receives $\nmax$ in the endowment $\finalendow$ (i.e; $\finaldata_i = \nmax$), it is impossible to find a Pareto-improvement $\finalendow^*$ of $\finalendow$, as it would require $\finaldata_i^* > \nmax$ which is impossible.

Let $\finalendow \in [0, \nmax]^m $ be a final endowment and $\finalendow^* \in [0, \nmax]^m$ a Pareto-improvement over $\finalendow$, that is:
\begin{align*}
    u_i(\finalendow^*) \geq u_i(\finalendow) \iff \frac{P(\finaldata^*_i)^{r+1}}{\sum_{j \in \firms} P(\finaldata^*_j)^r + \Poutside}  \geq \frac{P(\finaldata_i)^{r+1}}{\sum_{j \in \firms} P(\finaldata_j)^r + \Poutside} \quad \text{for all } i \in\firms,
\end{align*}
with at least one strict inequality. Denote 
$    C = \{i\in\firms, \finaldata_i^* > \finaldata_i  \} \quad \text{and} \quad \overline{C} = \firms \setminus C = \{i\in\firms, \finaldata_i^* \leq \finaldata_i  \}$. $C$ is necessarily non-empty. Indeed, if $C = \emptyset$, then $\finaldata_i^* \leq \finaldata_i \implies P(\finaldata_i^*)^r \leq P(\finaldata_i)^r$ for all $i \in\firms$ (as $P$ is strictly increasing and $r>0$), which implies that the sum of market shares is decreasing when going from $\finalendow$ to $\finalendow^*$:
\begin{align*}
    \sum_{i\in\firms} m_i(\finalendow^*) = \sum_{i\in\firms} \frac{P(\finaldata^*_i)^{r}}{\sum_{j \in \firms} P(\finaldata^*_j)^r + \Poutside} &= 1 - \frac{\Poutside}{\sum_{j \in \firms} P(\finaldata^*_j)^r + \Poutside} \\
    &\leq 1 - \frac{\Poutside}{\sum_{j \in \firms} P(\finaldata_j)^r + \Poutside} = \sum_{i\in\firms} m_i(\finalendow).
\end{align*}
Two possible cases then might arise; firstly, it can be that $m_i(\finalendow^*) \leq m_i(\finalendow)$ for all $i\in\firms$, in which case $u_i(\finalendow^*) =  m_i(\finalendow^*) P(\finaldata_i^*) \leq m_i(\finalendow) P(\finaldata_i) = u_i(\finalendow)$ for all $i\in\firms$, which contradicts the fact that $\finalendow^*$ is a Pareto-improvement over $\finalendow$, as this would require $u_i(\finalendow^*) > u_i(\finalendow)$ for at least one $i\in\firms$. 

Secondly, if there exists an $i\in\firms$ such that $m_i(\finalendow^*) > m_i(\finalendow)$, then as $\sum_{i\in\firms} m_i(\finalendow^*) \leq \sum_{i\in\firms} m_i(\finalendow)$, it implies that there exists a $k\in\firms$ such that $m_i(\finalendow^*) < m_i(\finalendow)$. Therefore, the profit of firm $k$ would strictly decrease, as $u_k(\finalendow^*) =  m_k(\finalendow^*) P(\finaldata_k^*) < m_k(\finalendow) P(\finaldata_k) = u_k(\finalendow)$, which would again be a contradiction.

Therefore, we have shown that $C \neq \emptyset $. We will show by contradiction that $\overline{C} = \emptyset$; assume that $\overline{C} \neq \emptyset$. Denote $t_i := P(\finaldata_i)^r $ and $t^*_i := P(\finaldata^*_i)^r $ for all $i\in\firms$, $t_C = \sum_{i \in C} t_i, t^*_C = \sum_{i \in C} t^*_i $ and $t_{\bar{C}} = \sum_{i \in {\bar{C}}} t_i, t^*_{\bar{C}} = \sum_{i \in {\bar{C}}} t^*_i $; then the market shares of firm $i$ under endowments $\finalendow, \finalendow^*$ are given by:
\begin{align*}
    m_i(\finalendow) &= \frac{P(\finaldata_i)^{r}}{\sum_{j \in \firms} P(\finaldata_j)^r + \Poutside} = \frac{t_i}{t_C + t_{\bar{C}} + \Poutside }
    = m_i(\finalendow^*) = \frac{t^*_i}{t^*_C + t^*_{\bar{C}} + \Poutside }.
\end{align*}

As $r>0$ and $P$ is strictly increasing, $t^*_i = P(\finaldata^*_i)^r > P(\finaldata_i)^r = t_i  $ for all $i \in C$, and similarly $t^*_i \leq t_i$ for all $i \in \overline{C}$. Therefore, we have that $ t^*_C > t_C $ and $t^*_{\bar{C}} \leq t_{\bar{C}}$, which gives. Define the sum of market shares of firms $\overline{C}$ under $\finalendow$ and $\finalendow^*$ as:
\begin{align*}
    m_{\bar{C}} &:= \sum_{i\in {\bar{C}}} m_i(\finalendow) =\frac{t_{\bar{C}}}{t_C + t_{\bar{C}} + \Poutside } \quad \text{and} \quad m_{\bar{C}}^* := \sum_{i\in {\bar{C}}} m_i(\finalendow^*) = \frac{t_{\bar{C}}^*}{t_C^* + t_{\bar{C^*}} + \Poutside }
\end{align*}
We have that:
\begin{align*}
    m_{\bar{C}}^* = \frac{t^*_{\bar{C}}}{t^*_C + t^*_{\bar{C}} + \Poutside} < \frac{t^*_{\bar{C}}}{t_C + t^*_{\bar{C}} + \Poutside}  &= 1 - \frac{t_C + \Poutside}{t_C + t^*_{\bar{C}} + \Poutside}  \\
    &\leq 1 - \frac{t_C + \Poutside}{t_C + t_{\bar{C}} + \Poutside} = \frac{t_{\bar{C}}}{t_C + t_{\bar{C}} + \Poutside} = m_{\bar{C}}
\end{align*}
Therefore, we have:
\begin{align*}
    m_{\bar{C}}^* < m_{\bar{C}} &\iff \sum_{i\in {\bar{C}}} m_i(\finalendow^*) < \sum_{i\in {\bar{C}}} m_i(\finalendow) \iff \sum_{i\in {\bar{C}}} \big( m_i(\finalendow^*) - m_i(\finalendow) \big) < 0,
\end{align*}
which implies that there exists a firm $k\in \overline{C}$ whose market share is strictly decreasing when moving from $\finalendow$ to $\finalendow^*$, that is: $m_k(\finalendow^*) < m_k(\finalendow)$. Moreover,  $k\in \overline{C} \implies \finaldata_k^* \leq \finaldata_k \implies P(\finaldata_k^*) \leq P(\finaldata_k) $, which means that the profit of firm $k$ is strictly decreasing:
\begin{align*}
    u_k(\finalendow^*) = m_k(\finalendow^*) P(\finaldata^*_k) < m_k(\finalendow) P(\finaldata_k) = u_i(\finalendow),
\end{align*}
which contradicts the fact that $\finalendow^*$ is a Pareto-improvement over $\finalendow$. Therefore, we necessarily have that $\overline{C} = \emptyset$, which means that for Pareto-improvement $\finalendow^*$ of a final endowment $\finalendow$, we necessarily have that $\finaldata_i^* > \finaldata_i$ for all $i\in\firms$, which concludes the proof.
\end{proof}

\section{Proofs of Section \ref{sec:fairness_and_mechanisms}} \label{app:missing_proofs_B}

\begin{proof}[Proof of Proposition \ref{prop:set_fair_rules}]
Consider an initial endowment $\initendow$, and let $R(\initendow) = (R_1(\initendow), \dots, R_m(\initendow))$ be an equal-gain contract: we have:
\begin{align*}
	&\frac{u_1(\initendow + R(\initendow))}{u_1(\initendow)} = \dots \ = \frac{u_m(\initendow + R(\initendow))}{u_m(\initendow)} \\
    &\iff \frac{P(\initdata_1 + R_1(\initendow))^{r+1}}{\sum_{j \in \firms} P(\initdata_j+R_j(\initendow))^r + \Poutside} \times \frac{\sum_{j \in \firms} P(\initdata_j)^r + \Poutside}{P(\initdata_1)^{r+1}}  \\
    &\qquad \qquad = \dots = \frac{P(\initdata_m + R_m(\initendow))^{r+1}}{\sum_{j \in \firms} P(\initdata_j+R_j(\initendow))^r + \Poutside} \times \frac{\sum_{j \in \firms} P(\initdata_j)^r + \Poutside}{P(\initdata_m)^{r+1}} \\
    &\iff \left(\frac{P(\initdata_1 + R_1(\initendow))}{P(\initdata_1)}\right)^{r+1} = \dots = \left(\frac{P(\initdata_m + R_m(\initendow))}{P(\initdata_m)}\right)^{r+1} \\
    &\iff \frac{P(\initdata_1 + R_1(\initendow))}{P(\initdata_1)} = \dots = \frac{P(\initdata_m + R_m(\initendow))}{P(\initdata_m)}.
\end{align*}
Define $j^* = \argmax_j \initdata_j$; we then have, for all $i\in\firms$:
\begin{align*}
    P(\initdata_i + R_i(\initendow)) &=  P(\initdata_{j^*} + R_{j^*}(\initendow)) \frac{P(\initdata_i)}{P(\initdata_{j^*})} \\
    &\iff R_i(\initendow) = P^{-1}  \left( P(\initdata_{j^*} + R_{j^*}(\initendow)) \frac{P(\initdata_i)}{P(\initdata_{j^*})} \right) - \initdata_i
\end{align*}
Therefore, given $\initendow$, $ R(\initendow) = (R_1(\initendow), \dots, R_m(\initendow))  $ is entirely determined by $R_{j^*}(\initendow)$. As $\finaldata_{j^*} = \initdata_{j^*} + \DSdata_{j^*} $, firm $j^*$ cannot receive more than $\sum_{k\in F\setminus\{j^*\}} \initdata_k$ datapoints, and therefore $R_{j^*}(\initendow)$ can take any value in:
\begin{align*}
    \left[0, \sum_{k\in F\setminus\{j^*\}} \initdata_k \right] = \left\{ \gamma  \sum_{k\in F\setminus\{j^*\}} \initdata_k, \quad \gamma \in [0, 1] \right\}.
\end{align*}
We obtain that for any $R_{j^*}(\initendow)$, there exists a $\gamma\in [0, 1]$ such that $R_{j^*}(\initendow) = \gamma \sum_{k\in F\setminus\{j^*\}} \initdata_k  $. Moreover, any $\gamma\in [0, 1]$ yields an acceptable value for $R_{j^*}(\initendow)$. Therefore, equal-gain contract $R$ corresponds to a $\gamma\in [0, 1]$ and is given by:
\begin{align*}
    R_i(\initendow) = P^{-1}  \left( P\Big(\initdata_{j^*} + \gamma \sum_{k\in F\setminus\{j^*\}} \initdata_k \Big) \frac{P(\initdata_i)}{P(\initdata_{j^*})} \right) - \initdata_i,
\end{align*}
which concludes the proof.
\end{proof}

\begin{proof}[Proof of Theorem \ref{thm:properties_fair_rules}]

Let us first show that relative market shares remain constant under any contract $\gfairule^\gamma$. Consider an initial endowment $\initendow$, a $\gamma \in [0, 1]$ and two firms $i, j\in\firms$. We have:
\begin{align*}
    \frac{m_i(\initendow + \gfairule^{\gamma}(\initendow) )}{m_j(\initendow + \gfairule^{\gamma}(\initendow) )} &=  \frac{P(\initdata_i + \gfairule^{\gamma}_{i}(\initendow) )^r}{\sum_{k \in \firms} P(\initdata_k + \gfairule^{\gamma}_{k}(\initendow))^r + \Poutside} \times \frac{\sum_{k \in \firms} P(\initdata_k + \gfairule^{\gamma}_{k}(\initendow))^r + \Poutside}{P(\initdata_j + \gfairule^{\gamma}_{j}(\initendow)  )^r} \\
    &= \left(\frac{P(\initdata_i + \gfairule^{\gamma}_{i}(\initendow) )}{P(\initdata_j + \gfairule^{\gamma}_{j}(\initendow)  )}\right)^r.
\end{align*}
Moreover, replacing $\gfairule^{\gamma}_{i}(\initendow)$ by its definition, defining $j^* = \argmax_j \initdata_j$, we have:
\begin{align*}
    P(\initdata_i + \gfairule^{\gamma}_{i}(\initendow) ) &= P\left( P^{-1}  \left( P\Big(\initdata_{j^*} + \gamma \sum_{k\in F\setminus\{j^*\}} \initdata_k \Big) \frac{P(\initdata_i)}{P(\initdata_{j^*})} \right) \right) \\
    &= P\Big(\initdata_{j^*} + \gamma \sum_{k\in F\setminus\{j^*\}} \initdata_k \Big) \frac{P(\initdata_i)}{P(\initdata_{j^*})}.
\end{align*}
Doing the same for $\gfairule^{\gamma}_{j}(\initendow) )$ gives:
\begin{align*}
    \frac{P(\initdata_i + \gfairule^{\gamma}_{i}(\initendow) )}{P(\initdata_j + \gfairule^{\gamma}_{j}(\initendow) )} &= P\Big(\initdata_{j^*} + \gamma \sum_{k\in F\setminus\{j^*\}} \initdata_k \Big) \frac{P(\initdata_i)}{P(\initdata_{j^*})} \times \left( P\Big(\initdata_{j^*} + \gamma \sum_{k\in F\setminus\{j^*\}} \initdata_k \Big) \frac{P(\initdata_j)}{P(\initdata_{j^*})} \right)^{-1} \\
    &= \frac{P(\initdata_i)}{P(\initdata_j)}.
\end{align*}
On the other hand, we have:
\begin{align*}
    \frac{m_i(\initendow  )}{m_j(\initendow)} = \frac{P(\initdata_i )^r}{\sum_{k \in \firms} P(\initdata_k )^r + \Poutside} \times \frac{\sum_{k \in \firms} P(\initdata_k )^r + \Poutside}{P(\initdata_j)^r} = \left(\frac{P(\initdata_i )^r}{P(\initdata_j )}\right)^r,
\end{align*}
which concludes the proof that relative market shares are constant under $\gfairule^{\gamma}$.

We can now use this result to show that for profit of any firm is increasing in $\gamma$ under contracts $\gfairule^{\gamma}$. For a given initial endowment $\initendow$ and $\gamma \in [0, 1] $, under contract $\gfairule^{\gamma}$, the amount of data $\gfairule^{\gamma}_{i}$ received by firm $i$ is increasing with $\gamma$ for all $i\in\firms$, as $r>0$ and:
\begin{align*}
    \gfairule^{\gamma}_{i}(\initendow) = P^{-1} \left( P\Big( \initdata_{j^*} + \gamma \sum_{k \in\firms\setminus\{j^*\}} \initdata_k \Big)  \frac{P(\initdata_i)}{P(\initdata_{j^*})} \right) - \initdata_i, \quad \text{with } j^* = \argmax_j \initdata_j,
\end{align*}
and $P, P^{-1}$ are strictly increasing and $\sum_{k \in\firms\setminus\{j^*\}} \initdata_k > 0$. Therefore, the precision $P(\initdata_i + \gfairule^{\gamma}_{i}(\initendow) )$ of firm $i$ under $\gfairule^{\gamma}$ is strictly increasing with $\gamma$, for all $i\in\firms$. Consequently, the sum of market shares under $\gfairule^{\gamma}$ is increasing with $\gamma$, as:
\begin{align*}
    \sum_{j\in\firms} m_j(\initendow + \gfairule^{\gamma}(\initendow)) = \sum_{j\in\firms}\frac{P(\initdata_j + \gfairule^{\gamma}_{j}(\initendow) )^r}{\sum_{k \in \firms} P(\initdata_k + \gfairule^{\gamma}_{k}(\initendow) )^r + \Poutside} = 1 - \frac{\Poutside}{\sum_{j \in \firms} P(\initdata_j + \gfairule^{\gamma}_{j}(\initendow) )^r},
\end{align*}
which implies that $\sum_{j\in\firms} m_j(\initendow + \gfairule^{\gamma}(\initendow)) \geq \sum_{j\in\firms} m_j(\initendow)$.

Consider any firm $i\in\firms$; the market share of firm $i$ is increasing with $\gamma$. Indeed, as proven above, the relative market shares between firms stay constant, which gives:
\begin{align*}
    \sum_{j\in\firms} m_j(\initendow + \gfairule^{\gamma}(\initendow))& = \sum_{j\in\firms} m_i(\initendow + \gfairule^{\gamma}(\initendow)) \frac{m_j(\initendow)}{m_i(\initendow)} \\
    &\iff \frac{m_i(\initendow + \gfairule^{\gamma}(\initendow)) }{m_i(\initendow)} = \frac{\sum_{j\in\firms} m_j(\initendow + \gfairule^{\gamma}(\initendow))}{\sum_{j\in\firms} m_j(\initendow )} \geq 1 \\
    &\iff m_i(\initendow + \gfairule^{\gamma}(\initendow)) \geq m_i(\initendow).
\end{align*}

As every firm $i\in\firms$ sees its market share increases and its precision increase strictly with $\gamma$, it implies that the profit $u_i(\initendow + \gfairule^{\gamma}(\initendow)) =  m_i(\initendow + \gfairule^{\gamma}(\initendow)) P(\initdata_i + \gfairule^{\gamma}_{i}(\initendow) ) $ strictly increases with $\gamma$ for all firms under contract $\gfairule^{\gamma}$. Moreover, the profit of every firm is maximized at $\gamma=1$.

Finally, as the profit of every firm is greater under $\onefairule := L^1$ than under $L^{\gamma} $ for any $\gamma \in [0, 1)$, contract $\onefairule$ Pareto-dominates any contract $L^{\gamma} $ for $\gamma \in [0, 1)$, so no such contract can be Pareto-optimal. Moreover,
\begin{align*}
\initdata_{j^*} + \onefairule_{j^*}(\initendow) = P^{-1} \biggl( P\Big( \initdata_{j^*} + \hspace{-0.3cm}\sum_{k \in\firms\setminus\{j^*\}} \hspace{-0.2cm}\initdata_k \Big)  \cdot\frac{P(\initdata_{j^*})}{P(\initdata_{j^*})} \biggr) = \nmax,
\end{align*}

which show that $\onefairule$ is a Pareto-optimal contract by Theorem \ref{thm:paretoset}, and in particular, is the only Pareto-optimal contract among equal-gain contracts.
\end{proof}

In order to prove Theorem \ref{thm:strategyproofness}, we prove the following two technical small lemmas.
\medskip

\begin{lemma} \label{lem:base_prop_conc_P}
For any concave precision function $P \in \precisionset$, for any $\Delta > 0$, the function $x \longmapsto \frac{P(x+\Delta)}{P(x)}$ defined for $x>0$ is strictly decreasing.
\end{lemma}

\begin{proof}
Let $x_2 > x_1 > 0$. We want to show that:
\begin{align*}
    \frac{P(x_1+\Delta)}{P(x_1)} > \frac{P(x_2+\Delta)}{P(x_2)} \iff P(x_1+\Delta)P(x_2) > P(x_2+\Delta)P(x_1).
\end{align*}
Define $\alpha := \frac{x_2 - x_1}{x_2 - x_1 + \Delta} \in (0, 1)$. We can write $x_2$ and $x_1+\Delta$ as convex combinations of $x_1$ and $x_2 + \Delta$:
\begin{align*}
    x_2 &= (1-\alpha)x_1 + \alpha(x_2 + \Delta), \\
    x_1 + \Delta &= \alpha x_1 + (1-\alpha)(x_2 + \Delta).
\end{align*}
By concavity of $P$, we have:
\begin{align*}
    P(x_2) &\geq (1-\alpha)P(x_1) + \alpha P(x_2 + \Delta) \text{ and 
    }
    P(x_1 + \Delta) \geq \alpha P(x_1) + (1-\alpha)P(x_2 + \Delta).
\end{align*}
Multiplying these two inequalities together yields:
\begin{align}
    P(x_1 + \Delta) P(x_2) &\geq \Big[ \alpha P(x_1) + (1-\alpha)P(x_2 + \Delta) \Big] \Big[ (1-\alpha)P(x_1) + \alpha P(x_2 + \Delta) \Big] \notag \\
    &= \alpha(1-\alpha)P(x_1)^2 + \alpha(1-\alpha)P(x_2+\Delta)^2 + \big(\alpha^2 + (1-\alpha)^2\big) P(x_1)P(x_2+\Delta) \notag \\
    &= P(x_1)P(x_2+\Delta) + \alpha(1-\alpha)\big( P(x_2+\Delta) - P(x_1) \big)^2, \label{eq:ineq_proof_P_conc}
\end{align}
using $\alpha^2 + (1-\alpha)^2 = 1 - 2\alpha(1-\alpha)$ on the last line. Because $P$ is strictly increasing, $x_2 + \Delta > x_1 \implies P(x_2+\Delta) > P(x_1)$. Moreover, $\alpha \in (0,1)$ implies $\alpha(1-\alpha) > 0$. Therefore:
\begin{align*}
    \alpha(1-\alpha)\big( P(x_2+\Delta) - P(x_1) \big)^2 > 0,
\end{align*}
which, when inputted in (\ref{eq:ineq_proof_P_conc}), yields:
\begin{align*}
    P(x_1 + \Delta)P(x_2) > P(x_1)P(x_2+\Delta) \iff \frac{P(x_1+\Delta)}{P(x_1)} > \frac{P(x_2+\Delta)}{P(x_2)}
\end{align*}
which concludes the proof.
\end{proof}

\begin{lemma} \label{lem:useful_ineq_P_conc}
For any concave precision function $P \in \precisionset$, for any $a, b \geq c \geq 0$ and $\delta \geq 0$, we have:
$$ P \left(a - c + P^{-1}\left( P(b+c+\delta) \frac{P(c)}{P(b)}   \right) \right) \leq P(b+c+\delta) \frac{P(a)}{P(b)} $$
\end{lemma}

\begin{proof}
Let $y = P^{-1}\left( P(b+c+\delta) \frac{P(c)}{P(b)}   \right)$, which implies 
\begin{align*}
    P(y) = P(b+c+\delta) \frac{P(c)}{P(b)} \iff  \frac{P(y)}{P(c)} =  \frac{P(b+c+\delta)}{P(b)}.
\end{align*}
As $c, \delta \geq 0$ and $P$ is increasing, $P(b+c+\delta ) \geq P(b) \implies P(y) / P(c) \geq 1 \implies y \geq c$. Moreover, substituting $P(b+c+\delta) = P(y) \frac{P(b)}{P(c)}$ in our original inequality gives:
\begin{align*}
    P \left(a - c + P^{-1}\left( P(b+c+\delta) \frac{P(c)}{P(b)}   \right) \right) &\leq P(b+c+\delta) \frac{P(a)}{P(b)} \iff P(a-c+y) \leq P(y) \frac{P(a)}{P(c)} \\
    &\iff \frac{P(y+a-c)}{P(y)} \leq \frac{P(c+a-c)}{P(c)},
\end{align*}

which is always true by Lemma \ref{lem:base_prop_conc_P}, as $\Delta = a-c \geq 0$ and $y \geq c$.
\end{proof}

\begin{proof}[Proof of Theorem \ref{thm:strategyproofness}]

Consider a concave precision function $P\in\precisionset$, an initial endowment $\initendow$, a firm $i\in\firms$, a competitor declaration $\decdata_{-i} $ and a potential lie $\decdata_i \in [0, \initdata_i]$. We want to prove that:
\begin{align*}
    u_i\big( \initendow + \onefairule( \initdata_i, \decdata_{-i}) \big) \geq u_i\big( \initendow + \onefairule( \decdata_i, \decdata_{-i}) \big).
\end{align*}
Denote $j^* = \argmax_{j\in\firms\setminus\{i\}} \decdata_j $ the index of the firm declaring the largest amount, other than firm $i$. We distinguish three cases:
\begin{enumerate}
    \item $\initdata_i \geq \decdata_{j^*} $ and $\decdata_i \geq \decdata_{j^*} $; firm $i$ always declares the most data, whether they are lying or being truthful.
    \item $\initdata_i \geq \decdata_{j^*} $ and $\decdata_i < \decdata_{j^*} $; firm $i$ declares the most data when being truthful, but not when lying.
    \item $\initdata_i < \decdata_{j^*} $ and $\decdata_i < \decdata_{j^*} $; firm $i$ never declares the most data, even when being truthful.
\end{enumerate}
For each of these three cases, we will compare the profit of firm $i$ when declaring truthfully $\initdata_i$ to its profit when lying and declaring $\decdata_i$. We will always denote $A=(A_1, \dots, A_m) := \onefairule( \initdata_i, \decdata_{-i}) $ the data received by all firms when firm $i$ is being truthful, and $B=(B_1, \dots, B_m) := \onefairule( \decdata_i, \decdata_{-i})$ the data received by all firms when firm $i$ lies. Denoting $\finalendow_A := \initendow + A$ the final endowment when firm $i$ is truthful and $\finalendow_B := \initendow + B$ the final endowment when firm $i$ lies, we want to prove that in each of these three cases, $u_i(\finalendow_A) \geq u_i(\finalendow_B)$. 

\medskip
\noindent\textbf{Case 1}: If $\initdata_i \geq \decdata_{j^*} $ and $\decdata_i \geq \decdata_{j^*} $:
\vspace{0.5em}

In this case, $i = \argmax_{j\in\firms} \decdata_j $, and thus when firm $i$ does not lie, for all $j\in\firms\setminus\{i\}$ firm $j$ receives:
\begin{align*}
    A_j = \onefairule_j(\initdata_i, \decdata_{-i}) = P^{-1} \left( P\Big( \initdata_i + \sum_{k \neq i} \decdata_k \Big)  \frac{P(\decdata_j)}{P(\initdata_i)} \right) - \decdata_j,
\end{align*}
and firm $i$ receives $A_i = \sum_{k\in\firms\setminus\{i\}} \decdata_k $. On the other hand, when firm $i$ lies, for all $j\in\firms$ firm $j$ receives:
\begin{align*}
    B_j = H_j(\decdata_i, \decdata_{-i}) = P^{-1} \left( P\Big( \sum_{k \in \firms} \decdata_k \Big)  \frac{P(\decdata_j)}{P(\decdata_i)} \right) - \decdata_j,
\end{align*}
and in particular firm $i$ receives $B_i = \sum_{k\in\firms\setminus\{i\}} \decdata_k = A_i$. As $A_i = B_i$ and the profit of firm $i$ is decreasing in the final endowment $N_j$ of firm $j$ for all $j\in\firms\setminus\{i\}$ (by Proposition \ref{prop:profit_monotonicity}), it suffices to prove that $B_j \geq A_j $ for all such $j$:
\begin{align*}
    B_j \geq A_j &\iff P^{-1} \left( P\Big( \sum_{k \in \firms} \decdata_k \Big)  \frac{P(\decdata_j)}{P(\decdata_i)} \right) - \decdata_j \geq P^{-1} \left( P\Big( \initdata_i + \sum_{k \in\firms\setminus\{i\}} \decdata_k \Big)  \frac{P(\decdata_j)}{P(\initdata_i)} \right) - \decdata_j \\
    &\iff \frac{P\Big( \decdata_i + \sum_{k \neq i} \decdata_k \Big)}{P(\decdata_i)} \geq   \frac{P\Big( \initdata_i + \sum_{k \in\firms\setminus\{i\}} \decdata_k \Big)}{P(\initdata_i)},
\end{align*}
which always holds from Lemma \ref{lem:base_prop_conc_P}, as $ \initdata_i \geq \decdata_i \geq 0$. Therefore, $\finalendow_{B, j} \geq \finalendow_{A, j}  $ and $\finalendow_{B, i} = \finalendow_{A, i}$, so by Proposition \ref{prop:profit_monotonicity}, we have that $u_i(\finalendow_A) \geq u_i(\finalendow_B)$.

\medskip
\noindent\textbf{Case 2}: If $\initdata_i \geq \decdata_{j^*} $ and $\decdata_i < \decdata_{j^*} $:
\vspace{0.5em}

In this case, when firm $i$ does not lie it is still declaring the largest amount of data, and thus for all $j\in\firms\setminus\{i\}$ firm $j$ receives:
\begin{align*}
    A_j = H_j(\initdata_i, \decdata_{-i}) = P^{-1} \left( P\Big( \initdata_i + \sum_{k \in \firms \setminus \{i\}} \decdata_k \Big)  \frac{P(\decdata_j)}{P(\initdata_i)} \right) - \decdata_j,
\end{align*}
as in the first case of the proof, and firm $i$ receives $A_i = \sum_{k \in \firms \setminus \{i\}} \decdata_k $. However, when firm $i$ lies, it is not the one declaring the largest amount of data anymore, and the formula for $B$ changes; for all $j\in\firms$, firm $j$ receives:
\begin{align*}
    B_j = H_j(\decdata_i, \decdata_{-i}) = P^{-1} \left( P\Big( \sum_{k\in\firms} \decdata_k \Big)  \frac{P(\decdata_j)}{P(\decdata_{j^*})} \right) - \decdata_j.
\end{align*}
In particular, firm $i$ receives:
\begin{align*}
    B_i = H_i(\decdata_i, \decdata_{-i}) = P^{-1} \left( P\Big( \sum_{k\in\firms} \decdata_k \Big)  \frac{P(\decdata_i)}{P(\decdata_{j^*})} \right) - \decdata_i
\end{align*}

In this case, $A_i \neq B_i$, so the previous method does not work anymore. To prove that $u_i(\initendow + A) \geq u_i(\initendow + B)$, we will build a curve $\mathcal{C}$ parametrized by $\gamma \in [0, 1] $ (using the $\gamma$-fair rules), and we will prove that the profit of firm $i$ is increasing with $\gamma$ along this curve. Moreover, point $A$ will correspond to the point on $\mathcal{C}$ where $\gamma = 1$, and will therefore maximize $u_i$ along $\mathcal{C}$. Finally, we will construct an intermediary point $C\in\mathcal{C}$ such that $B_i = C_i$, and we will then prove that $B_j \geq C_j$ for all $j\in\firms\setminus\{i\}$, which as earlier will imply that $u_i(\initendow + C) \geq u_i(\initendow + B)$. This will complete the proof, as we will already have proven that $u_i(\initendow + A) \geq u_i(\initendow + C)$ by construction of $\mathcal{C}$.

Let us first define $\mathcal{C}$ as the set, parametrized by $\gamma\in[0, 1]$, of the data received $\DSendow$ under the $\gamma$-fair rules $\gfairule^{\gamma}$, when firm $i$ declares $\initdata_i$ truthfully and all other firms declare $\decdata_{-i}$:
\begin{align*}
    \mathcal{C} := \left\{ C_\gamma = L^{\gamma}(\initdata_i, \decdata_{-i} ), \gamma \in [0, 1]  \right\},
\end{align*}
which gives, for any $j \in \firms \setminus \{i\}$:
\begin{align*}
    C^{\gamma}_{j} &= P^{-1} \left( P\Big( \initdata_i + \gamma \sum_{k \in \firms \setminus \{i\}} \decdata_k \Big)  \frac{P(\decdata_j)}{P(\initdata_i)} \right) - \decdata_j,
\end{align*}
and $C^{\gamma}_{i} = \gamma \sum_{k \in \firms \setminus \{i\}} \decdata_k$. We first want to prove that the profit $u_i(\initendow + C^{\gamma}) $ of firm $i$ is increasing with $\gamma$ along $\mathcal{C}$:
\begin{align*}
    u_i(\initendow + C^{\gamma}) &= \frac{P(\initdata_i + C^{\gamma}_{i})^{r+1}}{\sum_{j\in\firms} P(\initdata_j + C^{\gamma}_{j})^r + \Poutside}  = f(\gamma) \frac{f(\gamma)^r}{f(\gamma)^r + \sum_{j \in \firms \setminus \{i\}} g_j(\gamma)^r + \Poutside} \\
    &= f(\gamma)  \frac{1}{1 + \sum_{j \in \firms \setminus \{i\}} \big(\frac{g_j(\gamma)}{f(\gamma)}\big)^r + \big(\frac{\Poutside}{f(\gamma)^r}\big) }
\end{align*}

with $f:[0, 1] \rightarrow [0, 1] $ and $g_j:[0, 1] \rightarrow [0, 1]$ for all $j \in \firms \setminus \{i\}$ defined by:
\begin{align*}
    f(\gamma) &= P(\initdata_i + C^{\gamma}_{i}) = P\Big(\initdata_i + \gamma \sum_{k \in \firms \setminus \{i\}} \decdata_k \Big), \\
    g_j(\gamma) &= P(\initdata_j + C^{\gamma}_{j}) = P\left( \initdata_j - \decdata_j + P^{-1} \left( P\Big( \initdata_i + \gamma \sum_{k \in \firms \setminus \{i\}} \decdata_k \Big)  \frac{P(\decdata_j)}{P(\initdata_i)} \right)  \right)
\end{align*}

$f$ is strictly increasing with $\gamma$ as $P$ is strictly increasing and $\sum_{k \in \firms \setminus \{i\}} \decdata_k > 0$. Therefore, $\frac{\Poutside}{f(\gamma)^r}$ is decreasing with $\gamma$, and as $r>0$, showing that $\frac{g_j(\gamma)}{f(\gamma)}$ decreases with $\gamma$ for all $j \in \firms \setminus \{i\}$ would show that $u_i(\initendow + C^{\gamma})$ is increasing with $\gamma$.

Define $x := \initdata_i + \gamma \sum_{k \in \firms \setminus \{i\}} \decdata_k $ such that $f(\gamma) = P(x) $; $x$ is increasing with $\gamma$. Define \mbox{$ y := P^{-1} \left( P(x)  \frac{P(\decdata_j)}{P(\initdata_i)} \right)  $} such that $P(y) = P(x)  \frac{P(\decdata_j)}{P(\initdata_i)} $; $y$ is also increasing in $\gamma$, as $P, P^{-1} $ are strictly increasing. Therefore:
\begin{align*}
    \frac{g_j(\gamma)}{f(\gamma)} = \frac{P(y + \initdata_j - \decdata_j)}{P(x)} = \frac{P(y + \initdata_j - \decdata_j)}{P(y)} \frac{P(\decdata_j)}{P(\initdata_i)}
\end{align*}
As $\frac{P(\decdata_j)}{P(\initdata_i)}$ is constant with respect to $\gamma$ and $\initdata_j - \decdata_j \geq 0$, by Lemma \ref{lem:base_prop_conc_P} $\frac{g_j(\gamma)}{f(\gamma)}$ is decreasing with $y$ and thus with $\gamma$. This concludes that $u_i(\initendow + C^{\gamma})$ is increasing in $\gamma$. Moreover, note that point $A$ corresponds to the parameter $\gamma=1$: $A = C_1 \in \mathcal{C} $. Therefore, for any $\gamma \in[0, 1]$, we have that:
\begin{align}\label{eq:A_max_on_mathcalC}
    u_i(\initendow + C^{\gamma}) \leq u_i(\initendow + A).
\end{align}

Let us now search for an $\overline{\gamma}\in [0, 1]$ such that $B_i = C^{\overline{\gamma}}_{i} $:
\begin{align*}
    B_i = C^{\overline{\gamma}}_{i} &\iff P^{-1} \left( P\Big( \sum_{k\in\firms} \decdata_k \Big)  \frac{P(\decdata_i)}{P(\decdata_{j^*})} \right) - \decdata_i = \overline{\gamma}  \sum_{k \in \firms \setminus \{i\}} \decdata_k \\
    &\iff \overline{\gamma} = \frac{1}{\sum_{k \in \firms \setminus \{i\}} \decdata_k} \left[ P^{-1} \left( P\Big( \sum_{k\in\firms} \decdata_k \Big)  \frac{P(\decdata_i)}{P(\decdata_{j^*})} \right) - \decdata_i  \right].
\end{align*}
By construction, $\frac{P(\decdata_i)}{P(\decdata_{j^*})} \leq 1$ and $\frac{P\big( \sum_{k\in\firms} \decdata_k \big)}{P(\decdata_{j^*})} \geq 1$ so we have:
\begin{align*}
    P^{-1} ( P(\decdata_i )) \leq  P^{-1} \left( P\big( \sum_{k\in\firms} \decdata_k \big)  \frac{P(\decdata_i)}{P(\decdata_{j^*})} \right) \leq P^{-1} \Big( P\big( \sum_{k\in\firms} \decdata_k \big)  \Big) \iff 0 \leq B_i \leq \sum_{k \in \firms \setminus \{i\}} \decdata_k,
\end{align*}
and therefore $\overline{\gamma} \in [0, 1]$ and $C_{\overline{\gamma}} \in \mathcal{C} $ is well defined.

Let us now prove that for all $j \in \firms\setminus\{i\} $, $B_j \geq C^{\overline{\gamma}}_{j}$:
\begin{align}
    B_j \geq C^{\overline{\gamma}}_{j} &\iff P^{-1} \left( P\Big( \sum_{k\in\firms} \decdata_k \Big)  \frac{P(\decdata_j)}{P(\decdata_{j^*})} \right) - \decdata_j \geq P^{-1} \left( P\Big( \initdata_i + \overline{\gamma} \sum_{k \in \firms \setminus \{i\}} \decdata_k \Big)  \frac{P(\decdata_j)}{P(\initdata_{i})} \right) - \decdata_j \notag \\
    &\iff   \frac{P\Big( \sum_{k\in\firms} \decdata_k \Big)}{P(\decdata_{j^*})} \geq   \frac{P\Big( \initdata_{i} + \overline{\gamma} \sum_{k \in \firms \setminus \{i\}} \decdata_k \Big)}{P(\initdata_{i})} \notag \\
    &\iff P\Big( \initdata_{i} + \overline{\gamma} \sum_{k \in \firms \setminus \{i\}} \decdata_k \Big) \leq P\Big( \sum_{k\in\firms} \decdata_k \Big) \frac{P(\initdata_i)}{P(\decdata_{j^*})} \notag \\
    &\iff P\Bigg( \initdata_i - \decdata_i + P^{-1} \left( P\Big(\sum_{k\in\firms} \decdata_k \Big)  \frac{P(\decdata_i)}{P(\decdata_{j^*})} \right) \Bigg) \leq P\Big( \sum_{k\in\firms} \decdata_k \Big) \frac{P(\initdata_i)}{P(\decdata_{j^*})} \label{eq:Bj_leq_Cgammaj}
\end{align}
where we replace $\overline{\gamma}$ by its value in the last line. As we assumed that $\decdata_{j^*} > \decdata_i $ in this case, we can now simply apply Lemma \ref{lem:useful_ineq_P_conc} with $a=\initdata_i, b=\decdata_{j^*}, c=\decdata_i, \delta = \sum_{k \in \firms \setminus \{i, j^*\}} \decdata_k \geq 0 $ to conclude that Equation (\ref{eq:Bj_leq_Cgammaj}) always holds, and therefore that $B_j \geq C^{\overline{\gamma}}_{j}$ for all $j\in\firms\setminus\{i\}$.

In conclusion, as $B_i = C^{\overline{\gamma}}_{i} $ by construction of $\overline{\gamma}$, and as the profit of firm $i$ is decreasing in the final endowment $N_j$ of firm $j$ for all $j\in\firms\setminus\{i\}$ (by Proposition \ref{prop:profit_monotonicity}), we have that:
\begin{align*}
    u_i(\initendow + B) \leq u_i(\initendow +  C_{\overline{\gamma}}),
\end{align*}
which, combined with Equation (\ref{eq:A_max_on_mathcalC}), concludes that:
\begin{align*}
    u_i(\initendow + B) \leq u_i(\initendow +  A),
\end{align*}
as desired.

\medskip
\noindent\textbf{Case 3}: If $\initdata_i < \decdata_{j^*} $ and $\decdata_i < \decdata_{j^*} $:
\vspace{0.5em}

The logical reasoning of this case is very similar to the one in case 2; however, as we now assume that $\initdata_i < \decdata_{j^*} $, the set $\mathcal{C}$ and point $A$ will change. Nevertheless, the logic behind the proof stays exactly the same.

In this case, $\initdata_i < \decdata_{j^*} $, so even when being truthful firm $i$ will not be the firm declaring the largest amount of data. Thus, when firm $i$ does not lie, for all $j \in \firms\setminus\{i\}$ firm $j$ receives:
\begin{align*}
    A_j = H_j(\initdata_i, \decdata_{-i}) = P^{-1} \left( P\Big(\initdata_i + \sum_{k \in \firms \setminus \{i\}} \decdata_k \Big)  \frac{P(\decdata_j)}{P(\decdata_{j^*})} \right) - \decdata_j
\end{align*}
and firm $i$ receives:
\begin{align*}
    A_i = H_i(\initdata_i, \decdata_{-i}) = P^{-1} \left( P\Big(\initdata_i + \sum_{k \in \firms \setminus \{i\}} \decdata_k \Big)  \frac{P(\initdata_i)}{P(\decdata_{j^*})} \right) - \initdata_i.
\end{align*}
On the other hand, when firm $i$ lies, the situation is the same as in case 2, and the formula for $B$ does not change; for all $j\in\firms$, firm $j$ receives:
\begin{align*}
    B_j = H_j(\decdata_i, \decdata_{-i}) = P^{-1} \left( P\Big( \sum_{k\in\firms} \decdata_k \Big)  \frac{P(\decdata_j)}{P(\decdata_{j^*})} \right) - \decdata_j.
\end{align*}
In particular, firm $i$ receives:
\begin{align*}
    B_i = H_i(\decdata_i, \decdata_{-i}) = P^{-1} \left( P\Big( \sum_{k\in\firms} \decdata_k \Big)  \frac{P(\decdata_i)}{P(\decdata_{j^*})} \right) - \decdata_i
\end{align*}

Similarly as earlier, define $\mathcal{C}:= \left\{ C_\gamma = L^{\gamma}(\initdata_i, \decdata_{-i} ), \gamma \in [0, 1]  \right\}$, 
which gives:
\begin{align*}
    C^{\gamma}_{j} &= P^{-1} \left( P\Big( \decdata_{j^*} + \gamma \big( \initdata_i + \sum_{k \in \firms \setminus \{i, j^*\} } \decdata_k \big) \Big)  \frac{P(\decdata_j)}{P(\decdata_{j^*})} \right) - \decdata_j \quad \text{for all } j \in \firms \setminus \{i\}, \\
    C^{\gamma}_{i} &= P^{-1} \left( P\Big( \decdata_{j^*} + \gamma \big( \initdata_i + \sum_{k \in \firms \setminus \{i, j^*\} } \decdata_k \big) \Big)  \frac{P(\initdata_i)}{P(\decdata_{j^*})} \right) - \initdata_i.
\end{align*}

Let us first prove that, as earlier, $u_i(\initendow + C^{\gamma}) $ is increasing with $\gamma$ along $\mathcal{C}$:
\begin{align*}
    u_i(\initendow + C^{\gamma}) &= \frac{P(\initdata_i + C^{\gamma}_{i})^{r+1}}{\sum_{j\in\firms} P(\initdata_j + C^{\gamma}_{j})^r + \Poutside}  = f(\gamma)  \frac{1}{1 + \sum_{j \in \firms \setminus \{i\}} \big(\frac{g_j(\gamma)}{f(\gamma)}\big)^r + \big(\frac{\Poutside}{f(\gamma)^r}\big) },
\end{align*}
with $f$ and $g_j$ now defined by, for all $j \in \firms \setminus \{i\}$:
\begin{align*}
    f(\gamma) &:= P(\initdata_i + C^{\gamma}_{i}) = P\Big( \decdata_{j^*} + \gamma \big( \initdata_i + \sum_{k \in \firms \setminus \{i, j^*\} } \decdata_k \big) \Big)  \frac{P(\initdata_i)}{P(\decdata_{j^*})} \\
    g_j(\gamma) &:= P(\initdata_j + C^{\gamma}_{j}) = P\Bigg( \initdata_j - \decdata_j + P^{-1} \left( P\Big( \decdata_{j^*} + \gamma \big( \initdata_i + \sum_{k \in \firms \setminus \{i, j^*\} } \decdata_k \big) \Big)  \frac{P(\decdata_j)}{P(\decdata_{j^*})} \right)  \Bigg)
\end{align*}
As earlier, $f$ is strictly increasing in $\gamma$, so it is sufficient to show that $\frac{g_j(\gamma)}{f(\gamma)} $ is decreasing with $\gamma$ for all $j \in \firms \setminus \{i\}$. Define $x = \decdata_{j^*} + \gamma \big( \initdata_i + \sum_{k \in \firms \setminus \{i, j^*\} } \decdata_k \big)$ such that $f(\gamma) = P(x) \frac{P(\initdata_i)}{P(\decdata_{j^*})}  $; $x$ is increasing in $\gamma$. Define $y = P^{-1} \left( P(x) \frac{P(\decdata_j)}{P(\decdata_{j^*})} \right)$; $y$ is increasing with $x$ and therefore with $\gamma$, as $P^{-1}$ is increasing.  We have:
\begin{align*}
    P(y) = P(x) \frac{P(\decdata_j)}{P(\decdata_{j^*})} &\iff P(x) = P(y) \frac{P(\decdata_{j^*})}{P(\decdata_j)} \\
    &\iff f(\gamma) = P(y) \frac{P(\decdata_{j^*})}{P(\decdata_j)} \frac{P(\initdata_i)}{P(\decdata_{j^*})} = P(y) \frac{P(\initdata_i)}{P(\decdata_j)} 
\end{align*}
Putting everything together,we obtain:
\begin{align*}
    \frac{g_j(\gamma)}{f(\gamma)} = \frac{P(y + \initdata_j - \decdata_j)}{P(x)} = \frac{P(y + \initdata_j - \decdata_j)}{P(y) \frac{P(\initdata_i)}{P(\decdata_j)}} = \frac{P(y + \initdata_j - \decdata_j)}{P(y)} \frac{P(\decdata_j)}{P(\initdata_i)}
\end{align*}
which, as earlier, is decreasing in $y$ and therefore in $\gamma$, by Lemma \ref{lem:base_prop_conc_P}, for any $j \in \firms \setminus \{i\}$, which concludes that $u_i(\initendow + C^{\gamma})  $ is increasing with $\gamma$ along $\mathcal{C}$. Therefore, for any $\gamma \in[0, 1]$, we have that:
\begin{align*} 
    u_i(\initendow + C^{\gamma}) \leq u_i(\initendow + A). 
\end{align*}

Let us now find a $\overline{\gamma}\in[0, 1] $ such that $B_i = C^{\overline{\gamma}}_{i}$:
\begingroup
\allowdisplaybreaks
\begin{align}
    &B_i = C^{\overline{\gamma}}_{i} 
    \iff P^{-1} \left( P\Big(\decdata_i + \sum_{k \in \firms \setminus \{i\}} \decdata_k \Big)  \frac{P(\decdata_i)}{P(\decdata_{j^*})} \right) - \decdata_i \notag \\
    &\qquad \qquad\qquad\qquad\qquad = P^{-1} \left( P\Big( \decdata_{j^*} + \overline{\gamma} \big( \initdata_i + \sum_{k \in \firms \setminus \{i, j^*\} } \decdata_k \big) \Big)  \frac{P(\initdata_i)}{P(\decdata_{j^*})} \right) - \initdata_i \notag \\[5pt]
    &\iff P\Big( \decdata_{j^*} + \overline{\gamma} \big( \initdata_i + \sum_{k \in \firms \setminus \{i, j^*\} } \decdata_k \big) \Big)  = P\Bigg( \initdata_i - \decdata_i + P^{-1} \left( P\Big( \sum_{k\in \firms} \decdata_k \Big)  \frac{P(\decdata_i)}{P(\decdata_{j^*})} \right)  \Bigg) \frac{P(\decdata_{j^*})}{P(\initdata_i)} \label{eq:intermediary_step_def_overlinegamma} \\[5pt]
    &\iff \overline{\gamma} \big( \initdata_i + \sum_{k \in \firms \setminus \{i, j^*\} } \decdata_k \big) = P^{-1} \Bigg( P\Bigg( \initdata_i - \decdata_i + P^{-1} \left( P\Big(\sum_{k\in \firms} \decdata_k \Big)  \frac{P(\decdata_i)}{P(\decdata_{j^*})} \right)  \Bigg) \frac{P(\decdata_{j^*})}{P(\initdata_i)}  \Bigg) - \decdata_{j^*} \notag \\[5pt]
    &\iff \overline{\gamma} = \frac{1}{\initdata_i + \sum_{k \in \firms \setminus \{i, j^*\} } \decdata_k} \left[ P^{-1} \Bigg( P\Bigg( \initdata_i - \decdata_i + P^{-1} \left( P\Big(\sum_{k\in \firms} \decdata_k \Big)  \frac{P(\decdata_i)}{P(\decdata_{j^*})} \right)  \Bigg) \frac{P(\decdata_{j^*})}{P(\initdata_i)}  \Bigg) - \decdata_{j^*}   \right] \notag
\end{align}
\endgroup
As earlier, we need to check that this $\overline{\gamma} \in [0, 1] $ to verify that $C_{\overline{\gamma}}  $ is well defined. In Equation (\ref{eq:Bj_leq_Cgammaj}), we already showed that:
\begin{align*}
    P\Bigg( \initdata_i - \decdata_i + P^{-1} \left( P\Big(\sum_{k\in\firms} \decdata_k \Big)  \frac{P(\decdata_i)}{P(\decdata_{j^*})} \right) \leq P\Big( \sum_{k\in\firms} \decdata_k \Big) \frac{P(\initdata_i)}{P(\decdata_{j^*})},
\end{align*}
with the same necessary conditions $\initdata_i, \decdata_{j^*} \geq \decdata_i \geq 0 $. As $\initdata_i \geq \decdata_i$ and $P$ is increasing, $P\Big( \sum_{k\in\firms} \decdata_k \Big) \leq P\Big( \initdata_i + \sum_{k \in \firms \setminus \{i\}} \decdata_k \Big)$, which gives:
\begin{align*}
    P\Bigg( \initdata_i - \decdata_i + P^{-1} \left( P\Big(\sum_{k\in \firms} \decdata_k\Big)  \frac{P(\decdata_i)}{P(\decdata_{j^*})} \right)  \Bigg) \leq P\Big(\initdata_i + \sum_{k \in \firms \setminus \{i\}} \decdata_k \Big)  \frac{P(\initdata_i)}{P(\decdata_{j^*})}
\end{align*}

Moreover, $P(\decdata_i) \leq P\Big(\sum_{k\in \firms} \decdata_k\Big)  \frac{P(\decdata_i)}{P(\decdata_{j^*})} \iff \decdata_i \leq P^{-1} \Big( P\Big(\sum_{k\in \firms} \decdata_k\Big)  \frac{P(\decdata_i)}{P(\decdata_{j^*})} \Big) $. When putting both inequalities together, we obtain:
\begin{align*}
    & P(\initdata_i) \leq P\Bigg( \initdata_i - \decdata_i + P^{-1} \left( P\Big(\sum_{k\in \firms} \decdata_k\Big)  \frac{P(\decdata_i)}{P(\decdata_{j^*})} \right)  \Bigg) \leq P\Big(\initdata_i + \sum_{k \in \firms \setminus \{i\}} \decdata_k \Big)  \frac{P(\initdata_i)}{P(\decdata_{j^*})}  \\[5pt]
    &\iff P(\decdata_{j^*}) \leq P\Bigg( \initdata_i - \decdata_i + P^{-1} \left( P\Big(\sum_{k\in \firms} \decdata_k\Big)  \frac{P(\decdata_i)}{P(\decdata_{j^*})} \right)  \Bigg) \frac{P(\decdata_{j^*})}{P(\initdata_i)} \leq P\Big(\initdata_i + \sum_{k \in \firms \setminus \{i\}} \decdata_k \Big) \\[5pt]
    &\iff 0 \leq  P^{-1} \Bigg( P\Bigg( \initdata_i - \decdata_i + P^{-1} \left( P\Big(\sum_{k\in \firms} \decdata_k \Big)  \frac{P(\decdata_i)}{P(\decdata_{j^*})} \right)  \Bigg) \frac{P(\decdata_{j^*})}{P(\initdata_i)}  \Bigg) - \decdata_{j^*} \leq \initdata_i + \sum_{k \in \firms \setminus \{i\}} \decdata_k - \decdata_{j^*} \\[5pt]
    &\iff 0 \leq \overline{\gamma}  \leq 1
\end{align*}
Therefore, $C_{\overline{\gamma}} \in \mathcal{C} $ is well defined. Finally, we show that $B_j \geq C^{\overline{\gamma}}_{j} $ for all $j \in \firms \setminus \{i\}$:
\begin{align*}
    &B_j \geq C^{\overline{\gamma}}_{j}  \iff P^{-1} \left( P\Big(\sum_{k\in \firms} \decdata_k \Big)  \frac{P(\decdata_j)}{P(\decdata_{j^*})} \right) - \decdata_j \\
     &\qquad \qquad\qquad\qquad\qquad \geq P^{-1} \left( P\Big( \decdata_{j^*} + \overline{\gamma} \big( \initdata_i + \sum_{k \in \firms \setminus \{i, j^*\} } \decdata_k \big) \Big)  \frac{P(\decdata_j)}{P(\decdata_{j^*})} \right) - \decdata_j \\[5pt]
    &\iff P\Big(\sum_{k\in \firms} \decdata_k \Big)  \geq  P\Big( \decdata_{j^*} + \overline{\gamma} \big( \initdata_i + \sum_{k \in \firms \setminus \{i, j^*\} } \decdata_k \big) \Big)
\end{align*}
Reusing Equation (\ref{eq:intermediary_step_def_overlinegamma}), we obtain:
\begin{align*}
    &B_j \geq C^{\overline{\gamma}}_{j} \iff P\Big( \sum_{k \in \firms } \decdata_k \Big) \geq P\Bigg( \initdata_i - \decdata_i + P^{-1} \left( P\Big(\sum_{k\in \firms} \decdata_k \Big)  \frac{P(\decdata_i)}{P(\decdata_{j^*})} \right)  \Bigg) \frac{P(\decdata_{j^*})}{P(\initdata_i)} \\
    &\iff P\Bigg( \initdata_i - \decdata_i + P^{-1} \left( P\Big(\sum_{k\in \firms} \decdata_k \Big)  \frac{P(\decdata_i)}{P(\decdata_{j^*})} \right)  \Bigg) \leq  P\Big( \sum_{k \in \firms } \decdata_k \Big)  \frac{P(\initdata_i)}{P(\decdata_{j^*})} \iff (\ref{eq:Bj_leq_Cgammaj})
\end{align*}

which we already showed was always true for $\initdata_i \geq \decdata_i, \decdata_{j^*}\geq \decdata_i $.

\begin{align*}
    &\iff P\Big(\sum_{k\in F} \hat{n}_k \Big)  \ge  P\Big( \hat{n}_{j^*} + \overline{\gamma} \big( n_i + \sum_{k \in F \setminus \{i, j^*\} } \hat{n}_k \big) \Big) \\
    &\iff P\Big( \sum_{k \in F } \hat{n}_k \Big) \ge P\Bigg( n_i - \hat{n}_i + P^{-1} \left( P\Big(\sum_{k\in F} \hat{n}_k \Big)  \frac{P(\hat{n}_i)}{P(\hat{n}_{j^*})} \right)  \Bigg) \frac{P(\hat{n}_{j^*})}{P(n_i)} \\
    &\iff P\Bigg( n_i - \hat{n}_i + P^{-1} \left( P\Big(\sum_{k\in F} \hat{n}_k \Big)  \frac{P(\hat{n}_i)}{P(\hat{n}_{j^*})} \right)  \Bigg) \le  P\Big( \sum_{k \in F } \hat{n}_k \Big)  \frac{P(n_i)}{P(\hat{n}_{j^*})} \iff (3)
\end{align*}

We conclude in the exact same way as in case 2; as $B_i = C^{\overline{\gamma}}_{i} $ by construction of $\overline{\gamma}$, and as the profit of firm $i$ is decreasing in the final endowment $N_j$ of firm $j$ for all $j\in\firms\setminus\{i\}$ (by Proposition \ref{prop:profit_monotonicity}), we have that:
\begin{align*}
    u_i(\initendow + B) \leq u_i(\initendow +  C_{\overline{\gamma}}) \leq u_i(\initendow +  A),
\end{align*}
as desired.
\end{proof}

In order to prove Proposition \ref{prop:bound_gain_rounded_alloc}, we prove the following technical lemma.
\medskip

\begin{lemma}\label{lem:max_Ak_over_Bk}
For any sequences $(A_k)_{k \in \mathrm{F}}$ and $(B_k)_{k \in \mathrm{F}}$ strictly positive, with $\max_{k \in \mathrm{F}} \frac{A_k}{B_k} \ge 1$, and for any $P_{\varnothing} \ge 0$, it holds
\begin{align*}
\frac{\sum_{k \in \mathrm{F}} A_k + P_{\varnothing}}{\sum_{k \in \mathrm{F}} B_k + P_{\varnothing}} \le \max_{k \in \mathrm{F}} \frac{A_k}{B_k}.
\end{align*}
\end{lemma}

\begin{proof}
Let $M := \max_{k \in \mathrm{F}} \frac{A_k}{B_k} \ge 1$. Then $A_k \le M B_k$ for all $k \in \mathrm{F}$ and $P_{\varnothing} \le M P_{\varnothing}$, which gives $\sum_{k \in \mathrm{F}} A_k + P_{\varnothing} \le M \left(\sum_{k \in \mathrm{F}} B_k + P_{\varnothing}\right)$, from which the inequality follows.
\end{proof}

\begin{proof}[Proof of Proposition \ref{prop:bound_gain_rounded_alloc}]
    
To bound the relative utility gain from under-reporting under integer allocations, let $\finalendow^{\mathrm{tr}}$ and $\tilde{\finalendow}^{\mathrm{tr}}$ denote the continuous and rounded allocations under truthful reporting $\mathbf{n}$. Let $\finalendow^{\mathrm{lie}}$ and $\tilde{\finalendow}^{\mathrm{lie}}$ denote the allocations when firm $i$ deviates to $\hat{n}_i \le n_i$. Since the data-sharing contract can only strictly add data, all final dataset sizes satisfy $\finaldata_k, \tilde{\finaldata}_k \ge n_k \ge n_{\min}$ across all scenarios. By definition of any rounding rule, the maximum continuous-to-discrete deviation is bounded by 1: $|\tilde{\finaldata}_k - \finaldata_k| \le 1$ for any $k\in \firms$.

We decompose the ratio of rounded utilities via the continuous allocations:
\begin{equation}\label{eq:bridge}
\frac{u_i(\tilde{\finalendow}^{\mathrm{lie}})}{u_i(\tilde{\finalendow}^{\mathrm{tr}})} = \frac{u_i(\tilde{\finalendow}^{\mathrm{lie}})}{u_i(\finalendow^{\mathrm{lie}})}\cdot \frac{u_i(\finalendow^{\mathrm{lie}})}{u_i(\finalendow^{\mathrm{tr}})} \cdot \frac{u_i(\finalendow^{\mathrm{tr}})}{u_i(\tilde{\finalendow}^{\mathrm{tr}})}.
\end{equation}

By Theorem 5, which showed strategyproofness under continuous allocations, the middle term of Equation \ref{eq:bridge} satisfies $\frac{u_i(\finalendow^{\mathrm{lie}})}{u_i(\finalendow^{\mathrm{tr}})} \le 1$.

For the first term of Equation \ref{eq:bridge}, replacing by the definition of $u_i$ yields:
\begin{align*}
\frac{u_i(\tilde{\finalendow}^{\mathrm{lie}})}{u_i(\finalendow^{\mathrm{lie}})} = \left( \frac{P(\tilde{\finaldata}_i^{\mathrm{lie}})}{P(\finaldata_i^{\mathrm{lie}})} \right)^{r+1} \frac{\sum_{k \in \mathrm{F}} P(\finaldata_k^{\mathrm{lie}})^r + P_{\varnothing}}{\sum_{k \in \mathrm{F}} P(\tilde{\finaldata}_k^{\mathrm{lie}})^r + P_{\varnothing}}.
\end{align*}

Since $P$ is non-decreasing and using that $|\tilde{\finaldata}_k - \finaldata_k| \le 1$ for all $k\in\firms$, we can upper-bound the numerator using $\tilde{\finaldata}_i^{\mathrm{lie}} \le \finaldata_i^{\mathrm{lie}} + 1$ and lower-bound the denominator using $\tilde{\finaldata}_k^{\mathrm{lie}} \ge \finaldata_k^{\mathrm{lie}} - 1$ for all $k \in \firms$, which yields:
\begin{align*}
\frac{u_i(\tilde{\finalendow}^{\mathrm{lie}})}{u_i(\finalendow^{\mathrm{lie}})} \le \left( \frac{P(\finaldata_i^{\mathrm{lie}} + 1)}{P(\finaldata_i^{\mathrm{lie}})} \right)^{r+1} \frac{\sum_{k \in \mathrm{F}} P(\finaldata_k^{\mathrm{lie}})^r + P_{\varnothing}}{\sum_{k \in \mathrm{F}} P(\finaldata_k^{\mathrm{lie}} - 1)^r + P_{\varnothing}}.
\end{align*}

We bound the two parts of this product independently.  As $\finaldata_i^{\mathrm{lie}} > n_{\min} - 1$, applying Lemma \ref{lem:base_prop_conc_P} yields $
\frac{P(\finaldata_i^{\mathrm{lie}} + 1)}{P(\finaldata_i^{\mathrm{lie}})}  \le \frac{P(n_{\min})}{P(n_{\min} - 1)}$. 
Moreover, applying Lemma \ref{lem:max_Ak_over_Bk} with $A_k = P(\finaldata_k^{\mathrm{lie}})^r$ and $B_k = P(\finaldata_k^{\mathrm{lie}} - 1)^r$ and Lemma \ref{lem:base_prop_conc_P} using $\finaldata_k^{\mathrm{lie}} - 1 \ge n_{\min} - 1$, we obtain:
\begin{align*}
\frac{\sum_{k \in \mathrm{F}} P(\finaldata_k^{\mathrm{lie}})^r + P_{\varnothing}}{\sum_{k \in \mathrm{F}} P(\finaldata_k^{\mathrm{lie}} - 1)^r + P_{\varnothing}} \le \max_{k \in \mathrm{F}} \left( \frac{P(\finaldata_k^{\mathrm{lie}})}{P(\finaldata_k^{\mathrm{lie}} - 1)} \right)^r \le \left( \frac{P(n_{\min})}{P(n_{\min}-1)} \right)^r.
\end{align*}

Combining these two upper bounds yields:
\begin{align*}
\frac{u_i(\tilde{\finalendow}^{\mathrm{lie}})}{u_i(\finalendow^{\mathrm{lie}})} \leq \left( \frac{P(n_{\min})}{P(n_{\min}-1)} \right)^{2r+1}.
\end{align*}

Symmetrically, for the third term of Equation \ref{eq:bridge},
\begin{align*}
\frac{u_i(\finalendow^{\mathrm{tr}})}{u_i(\tilde{\finalendow}^{\mathrm{tr}})} = \left( \frac{P(\finaldata_i^{\mathrm{tr}})}{P(\tilde{\finaldata}_i^{\mathrm{tr}})} \right)^{r+1} \frac{\sum_{k \in \mathrm{F}} P(\tilde{\finaldata}_k^{\mathrm{tr}})^r + P_{\varnothing}}{\sum_{k \in \mathrm{F}} P(\finaldata_k^{\mathrm{tr}})^r + P_{\varnothing}},
\end{align*}
we upper-bound the numerator using $\finaldata_i^{\mathrm{tr}} \le \tilde{\finaldata}_i^{\mathrm{tr}} + 1$ and lower-bound the denominator using $\finaldata_k^{\mathrm{tr}} \ge \tilde{\finaldata}_k^{\mathrm{tr}} - 1$ for all $k \in \mathrm{F}$. Since $\tilde{\finaldata}_k^{\mathrm{tr}} \ge n_{\min}$ for all $k \in \mathrm{F}$, applying the same logic as for the first term, we obtain:
\begin{align*}
\frac{u_i(\finalendow^{\mathrm{tr}})}{u_i(\tilde{\finalendow}^{\mathrm{tr}})} \le \left( \frac{P(n_{\min})}{P(n_{\min}-1)} \right)^{2r+1}.
\end{align*}

Substituting these upper bounds into Equation \ref{eq:bridge} yields:
\begin{align*}
\frac{u_i(\tilde{\finalendow}^{\mathrm{lie}})}{u_i(\tilde{\finalendow}^{\mathrm{tr}})} \le \left( \frac{P(n_{\min})}{P(n_{\min}-1)} \right)^{2r+1} \cdot 1 \cdot \left( \frac{P(n_{\min})}{P(n_{\min}-1)} \right)^{2r+1} = \left( \frac{P(n_{\min})}{P(n_{\min} - 1)} \right)^{4r + 2},
\end{align*}
which concludes the proof.
\end{proof}

\section{Concavity Test} \label{app:discussion_sec_experiments}

To test the concavity of the empirical precision functions in Figure \ref{fig:multiple_P_curves}, we analyzed whether the secant slopes decreases, which would characterize a concave function. For this, at each sampled point of a curve, we calculated the secant slope 
$$m_i = \frac{y_i - y_{i-s}}{x_i - x_{i-s}}, $$
for a step size $s \in \{1, 2, 3\} $, and calculated the percentage of points for which $m_i \leq m_{i-s} $. 

\begin{table}[ht]
\small
\centering
\caption{Percentage of strictly decreasing adjacent secant slopes across window sizes $s$.}
\label{tab:empirical_concavity}
\begin{tabular}{l|ccc}
\hline
Menu Size & Step $s = 1$ & Step $s = 2$ & Step $s = 3$ \\
\hline
$k = 1$ & 66.67\% (12/18) & 87.50\% (14/16) & 100.00\% (14/14) \\
$k = 3$ & 72.22\% (13/18) & 87.50\% (14/16) & 100.00\% (14/14) \\
$k = 5$ & 83.33\% (15/18) & 100.00\% (16/16) & 100.00\% (14/14) \\
\hline
\end{tabular}
\end{table}

The results are displayed in Table \ref{tab:empirical_concavity}. For a step-size of $s=1$, the secant slopes are not always decreasing, which implies that the functions are not entirely concave at a micro level. This can mostly be attributed to changes in hyperparameters, as the eye-test on Figure \ref{fig:multiple_P_curves} does indeed show that when two consecutive points have different hyperparameters (denoted by the change between squares, triangles, and circles), the interpolated precision function can sometimes appear not to be entirely concave. We expect that performing a more thorough hyperparameter selection would lead to smoother transitions between hyperparameter regimes, reducing the non-concave behavior in these transition zones. That said, for larger windows, the secant slopes become all decreasing for all fitted precision functions, which shows that, at a macro level, the functions display a generally concave behavior.

\end{document}